\documentclass[twoside, letterpaper]{article}
\pdfoutput=1
\usepackage[hidelinks]{hyperref}
\usepackage[sectionbib]{natbib}
\usepackage{chapterbib}
\usepackage[accepted]{aistats2015}
\usepackage{amsmath,amssymb, theorem, stmaryrd,enumerate,bbm,bm}
\usepackage{booktabs}
\usepackage{multirow}
\usepackage{graphicx}
\usepackage{subfig}
\setcitestyle{square}
\usepackage{algorithm}
\usepackage{algpseudocode}

\newcommand{\g}{\,|\,}

\newcommand{\tmdate}[1]{\today}

{\theorembodyfont{\rmfamily}}
\newcommand{\tmfloatcontents}{}
\newlength{\tmfloatwidth}
\newcommand{\tmfloat}[5]{
  \renewcommand{\tmfloatcontents}{#4}
  \setlength{\tmfloatwidth}{\widthof{\tmfloatcontents}+1in}
  \ifthenelse{\equal{#2}{small}}
    {\ifthenelse{\lengthtest{\tmfloatwidth > \linewidth}}
      {\setlength{\tmfloatwidth}{\linewidth}}{}}
    {\setlength{\tmfloatwidth}{\linewidth}}
  \begin{minipage}[#1]{\tmfloatwidth}
    \begin{center}
      \tmfloatcontents
      \captionof{#3}{#5}
    \end{center}
  \end{minipage}}

\begin{document}

\twocolumn[

\aistatstitle{The Bayesian Echo Chamber: Modeling Social Influence via
  Linguistic Accommodation}

\aistatsauthor{Fangjian Guo \And Charles Blundell \And Hanna Wallach
  \And Katherine Heller}

\aistatsaddress{ Duke University\\Durham, NC,
  USA\\ {\texttt{guo@cs.duke.edu}} \And Gatsby Unit, UCL\\London, UK\\ {\texttt{c.blundell@gatsby.ucl.ac.uk~~}} \And
  Microsoft Research\\New York, NY, USA\\ {\texttt{~~wallach@microsoft.com}} \And Duke
  University\\Durham, NC, USA\\ {\texttt{kheller@stat.duke.edu}}} ]

\begin{abstract}
We present the Bayesian Echo Chamber, a new Bayesian generative model
for social interaction data. By modeling the evolution of people's
language usage over time, this model discovers latent influence
relationships between them. Unlike previous work on inferring
influence, which has primarily focused on simple temporal dynamics
evidenced via turn-taking behavior, our model captures more nuanced
influence relationships, evidenced via linguistic accommodation
patterns in interaction content. The model, which is based on a
discrete analog of the multivariate Hawkes process, permits a fully
Bayesian inference algorithm. We validate our model's ability to
discover latent influence patterns using transcripts of arguments
heard by the US Supreme Court and the movie ``12 Angry Men.'' We
showcase our model's capabilities by using it to infer latent
influence patterns from Federal Open Market Committee meeting
transcripts, demonstrating state-of-the-art performance at uncovering
social dynamics in group discussions.
\end{abstract}

\vspace{-0.2em}
\section{INTRODUCTION}

As increasing quantities of social interaction data become available,
often through online sources, researchers strive to find new ways of
using these data to learn about human behavior. Most social
processes, in which people or groups of people interact with one
another in order to achieve specific (and sometimes contradictory)
goals, are extremely complex. In order to construct realistic models
of these social processes, it is therefore necessary to take into account
their structure (e.g., who spoke with whom), content (e.g., what was
said), and temporal dynamics (e.g., when they spoke).

When studying social processes, one of the most pervasive questions is
``who influences whom?'' This question is of interest not only to
sociologists and psychologists, but also to political scientists,
organizational scientists, and marketing researchers. Since influence
relationships are seldom made explicit, they must be inferred from
other information. Influence has traditionally been
studied by analyzing declared structural links in observed networks,
such as Facebook ``friendships''~\citep{backstrom06group}, paper
citations~\citep{de_solla_price65networks}, and bill
co-sponsorships~\citep{fowler06legislative}. For many domains,
however, explicitly stated links do not exist, are unreliable, or fail
to reflect pertinent behavior. In these domains, researchers have
used observed interaction dynamics as a proxy by which to infer influence and
other social relationships. Much of this work has
concentrated (either implicitly or explicitly) on turn-taking behavior---i.e.,~``who acts next.''

In this paper, we take a different approach: we move beyond
turn-taking behavior, and present a new model, the Bayesian Echo
Chamber, that uses observed interaction content, in the context of temporal
dynamics, to capture influence. Our model draws upon a substantial body
of work within sociolinguistics indicating that when two people
interact, either orally or in writing, the use of a word by one person
can increase the other person's probability of subsequently using that
word. Furthermore, the extent of this increase depends on power differences and
influence relationships: the language used by a less powerful person
will drift further so as to more closely resemble or ``accommodate''
the language used by more powerful people. This phenomenon is known as
linguistic accommodation~\citep{west10introducing}. We demonstrate that
linguistic accommodation can reveal more nuanced influence patterns
than those revealed by simple reciprocal behaviors such as
turn-taking.

The Bayesian Echo Chamber is a new, mutually exciting, dynamic
language model that combines ideas from Hawkes
processes~\citep{Hawkes1971a}
with ideas from Bayesian language
modeling. We draw inspiration from Blundell et al.'s model of turn-taking
behavior~\citeyearpar{BluHelBec2012a} (described in section~\ref{sec:turn-taking}) to define a new model
of the mutual excitation of words in social interactions. This
approach, which leverages a discrete analog of a
multivariate Hawkes process, enables the Bayesian Echo Chamber to capture linguistic
accommodation patterns via latent influence variables. These variables
define a weighted influence network that reveals fine-grained
information about who influences whom.

We provide details of the Bayesian Echo Chamber in
section~\ref{sec:model}, including an MCMC algorithm for inferring the latent
influence variables (and other parameters) from real-world data. To
validate this algorithm, we provide parameter recovery results
obtained using synthetic data. In section~\ref{sec:experiments} we compare to several
baseline models on data sets including
arguments heard by the US Supreme Court~\citep{macwhinney2007talkbank}
and the transcript of the 1957 movie ``12 Angry Men.'' We compare influence networks inferred using our model to
those inferred using Blundell et al.'s model. We show that by focusing
on linguistic accommodation patterns, our model infers
different---more substantively meaningful---influence networks than those inferred
from turn-taking behavior.
We also combine our model with Blundell et al.'s so as to jointly model turn-taking and
linguistic accommodation. We investigate the possibility of tying the
latent influence parameters
to see if a single global notion of influence can be discovered.
Finally, we showcase our model's potential as an exploratory
analysis tool for social scientists using recently released
transcripts of Federal Reserve's Federal Open Market Committee meetings.

\vspace{-0.2em}
\section{INFLUENCE VIA TURN-TAKING}
\label{sec:turn-taking}

In this section, we give a brief description of a variant of
Blundell et al.'s
model for inferring influence from turn-taking behavior.
Unlike Blundell et al.'s original paper, which modeled pairwise actions, we
concentrate on a broadcast or group discussion setting
appropriate for the data that we wish to model. (We also do not cluster participants by their interaction patterns.) This setting, in which every
utterance is heard by and thus potentially influences every
participant, occurs in many scenarios of interest to social scientists. Furthermore,
the comparatively information-impoverished nature of this setting
makes it one in which ability to infer influence relationships is
deemed extremely valuable.

Blundell et al.'s model specifies a probabilistic generative process
for the time stamps $\mathcal{T} = \{ \mathcal{T}^{(p)} \}_{p=1}^P$ associated with
a set of actions made by $P$ people. In a group discussion setting,
these actions correspond to utterances, and the model captures who
will speak next and when that next utterance will occur. Letting
$N^{(p)}(T)$ denote the total number of utterances made by person $p$
over the entire observation interval $[0,T)$, each utterance made by
  $p$ is associated with a time stamp indicating its start time, i.e.,
  $\mathcal{T}^{(p)} = \{ t_n^{(p)} \}_{n=1}^{N^{(p)}(T)}$. We assume
  that the duration of each utterance $\Delta t_n^{(p)}$ is observed
  and that its end time ${t'}^{(p)}_n$ can be calculated from its
  start time and duration: ${t'}^{(p)}_n = t^{(p)}_n + \Delta
  t_n^{(p)}$.

Hawkes processes~\citep{Hawkes1971a}---a class of self- and mutually
exciting doubly stochastic point processes---form the mathematical
foundation of Blundell et al.'s model. A Hawkes process is a
particular form of
inhomogeneous Poisson process with a conditional stochastic
rate function $\lambda(t)$ that depends on the time stamps of all
events prior to time $t$.
Blundell et al.~model turn-taking interactions using coupled Hawkes
processes. For a group discussion setting, we instead define a
  multivariate Hawkes process, in which each person $p$ is
associated with his or her own Hawkes process defined on $(0, \infty)$. Letting $N^{(p)}(\cdot)$ denote the counting measure of person $p$'s
Hawkes process, which takes as its argument an interval $[a, b)$ and
  returns the number of utterances made by $p$ during that
  interval, the stochastic rate function for $p$'s
    Hawkes process is
\begin{align}
  \label{eq:rate}
    \lambda^{(p)}(t) &= \lambda_0^{(p)} +
    \sum_{q \neq p} \int_{0}^{t^-}
      g^{(qp)}(t, u)
          \,
            \textrm{d}N^{(q)}(u)\\
                &=
                  \lambda_0^{(p)} + \sum_{q \neq p} \sum_{n:
                      {t'}_n^{(q)} < t} g^{(qp)}(t, {t'}^{(q)}_n),
\end{align}
where $\lambda_0^{(p)}$ is person $p$'s base rate of utterances and
$g(t, u)$ is a non-negative stationary kernel
function that specifies the extent to which an event at time $u < t$
increases the instantaneous rate at time $t$, as well as the way in
which this increase decays over time. Person $p$'s
rate function is coupled with the Hawkes processes of the other $P-1$
people via their respective counting measures $\{N^{(q)}(\cdot) \}_{q
  \neq p}$ and the kernel function $g(t, u)$. Note that time stamp ${t'}_n^{(q)}$ is the \emph{end} time of the
$n^{\textrm{th}}$ utterance made by person $q$. Consequently, an
utterance made by person $q$ only causes an increase in $p$'s instantaneous
rate after $q$'s utterance is complete.

Blundell et al.~use a standard exponential kernel function of the form
$g^{(qp)}(t, u) = \nu^{(qp)}\exp{\left( - (t -
      u) \,/\, \tau_T^{(p)}\right)}$,
although the shape of the non-negative kernel function $g^{(qp)}(t, u)$ could instead
be
learned in a non-parametric fashion~\citep{zhou2013learning} if desired. Non-negative parameter $\nu^{(qp)}$ controls the degree of
instantaneous excitation from person $q$ to person $p$, while $\tau_T^{(p)}$ is a time decay parameter specific to person $p$
that characterizes how fast excitation decays. Since the goal is to
model influence between people, self-excitation is prohibited by
enforcing $\nu^{(pp)} = 0$.
Since a larger value of
$\nu^{(qp)}$ will result in a higher instantaneous rate of utterances
for person $p$, the non-negative parameters $\{\{ \nu^{(qp)} \}_{q \neq p}
\}_{p=1}^P$ define a weighted influence network that reflects
conversational
turn-taking behavior---i.e.,~who is likely to speak next and
when that next utterance will occur.
Details of an inference algorithm and appropriate priors for this
turn-taking-based model can be found in the supplementary material.



\vspace{-0.2em}
\section{INFLUENCE VIA LINGUISTIC ACCOMMODATION}
\label{sec:model}

In this section, we present our new dynamic Bayesian language
model, the Bayesian Echo Chamber. This model specifies a
probabilistic generative process for the words that occur in a set
of utterances $\{ \mathcal{W}^{(p)} \}_{p=1}^P$ made by $P$ people,
conditioned on the utterance start times and durations. Letting
$N^{(p)}(T)$ denote the total number of utterances made by person $p$
over the interval $[0,T)$, each utterance made by
  $p$ consists of $L_n^{(p)}$ word tokens, i.e., $\mathcal{W}^{(p)} = \{ \{
  w_{l,n}^{(p)} \}_{l=1}^{L_n^{(p)}} \}_{n=1}^{N^{(p)}(T)}$. Each
  token is an instance of one of $V$ unique word types.



The generative process for each token draws upon ideas from both
dynamic Bayesian language modeling and multivariate Hawkes
processes. The $l^{\textrm{th}}$ token in the $n^{\textrm{th}}$ utterance made by
person $p$ is drawn from categorical distribution specific to that
utterance:
$w_{l,n}^{(p)} \sim \textrm{Categorical}\,(\boldsymbol{\phi}_{n}^{(p)})$,
where $\boldsymbol{\phi}_n^{(p)}$ is a $V$-dimensional discrete probability
vector. Each such probability vector is in turn drawn from a Dirichlet
distribution with a person-specific concentration (or precision) parameter
and an utterance-specific base measure:
$\boldsymbol{\phi}_n^{(p)} \sim \text{Dirichlet}\,(\alpha^{(p)}, \boldsymbol{B}_n^{(p)})$.
Concentration parameter $\alpha^{(p)}$ is a positive scalar that
determines the variance of the distribution, while base measure
$\boldsymbol{B}_n^{(p)}$ is a $V$-dimensional discrete probability vector
that specifies the mean of the distribution and satisfies
 \begin{equation}
B^{(p)}_{v,n} \propto \beta^{(p)}_v + \sum_{q \neq p}
\rho^{(qp)} \psi^{(qp)}_{v, n} \ \textrm{and}\ \sum_{v=1}^{V}
B_{v,n}^{(p)}=1. \label{eqs:base-measure}
\end{equation}
$V$-dimensional vector $\boldsymbol{\beta}^{(p)} \in
\mathbb{R}_{+}^V$ characterizes person $p$'s inherent language
usage. Non-negative parameter $\rho^{(qp)}$ controls the degree of
linguistic excitation from person $q$ to person $p$. Self-excitation
is prohibited by enforcing $\rho^{(pp)} = 0$. Finally,
$\boldsymbol{\psi}^{(qp)}_n \in \mathbb{R}_{+}^V$ is a $V$-dimensional
vector of decayed excitation pseudocounts, constructed from all
utterances made by person $q$ prior to person $p$'s $n^{\textrm{th}}$
utterance, satisfying
\begin{align}
\psi^{(qp)}_{v,n} &=
  \sum_{m : {t'}^{(q)}_m < t^{(p)}_n}
      \left( \sum_{l=1}^{L_m^{(q)}} \mathbf{1}(w_{l,m}^{(q)} =
      v)\right) \times {} \notag\\
        &\quad\quad \exp{\left(-\frac{t_n^{(p)} -
                  {t'}_m^{(q)}}{\tau_L^{(p)}}\right)},
  \label{eq:lingrate}
\end{align}
where $\mathbf{1}(\cdot)$ is the indicator function.
The inner sum is therefore
equal to the number of tokens of type $v$ in person
$q$'s $m^{\textrm{th}}$ utterance. Note that ${t'}_m^{(q)}$ is the
  end time of that utterance. Consequently, an utterance made by
$q$ only affects
$\boldsymbol{\psi}^{(qp)}$, and hence base measure $\boldsymbol{B}_n^{(p)}$,
after $q$'s utterance is complete. Finally, $\tau_L^{(p)}$ is a time decay
parameter specific to person $p$ that characterizes how fast
excitation decays. Since a larger value of $\rho^{(qp)}$ will
increase the probability of person $p$ using word types previously
used by person $q$, the parameters $\{ \{ \rho^{(qp)} \}_{q \neq
  p}\}_{p=1}^P$ define a weighted influence network that
reflects linguistic accommodation.
A graphical model depicting the dependencies
between utterances is in the supplementary material.

\vspace{-0.2em}
\subsection{Inference}

For real-world group discussions, the utterance contents $\mathcal{W}
= \{ \mathcal{W}^{(p)}\}_{p=1}^P$, start times $\mathcal{T}$, and
durations ${\mathcal{D}}$ are observed, while parameters $\Theta = \{
\{ \boldsymbol{\phi}_n^{(p)} \}_{n=1}^{N^{(p)}(T)}, \alpha^{(p)},
\boldsymbol{\beta}^{(p)}, \{ \rho^{(qp)} \}_{q \neq p}, \tau_L^{(p)}
\}_{p=1}^P$ are unobserved; however, information about
the values of
these unobserved parameters can be quantified via their posterior distribution
given $\mathcal{W}$, $\mathcal{T}$, and
$\mathcal{D}$, i.e.,
$P(\Theta \g \mathcal{W}, \mathcal{T}, \mathcal{D}) \propto
  P(\mathcal{W} \g \Theta, \mathcal{T}, \mathcal{D})\,P(\Theta)$.
  The likelihood term can be factorized into the following product due
  to our model's independence assumptions:
\begin{align*}
  &P(\mathcal{W} \g \Theta, \mathcal{T}, \mathcal{D}) = \\
  &\quad\prod_{p=1}^P \prod_{n=1}^{N^{(p)}(T)} P(\boldsymbol{w}_n^{(p)} \g
       \{\{\boldsymbol{w}_m^{(q)}\}_{m : {t'}_m^{(q)}
       < t^{(p)}_n} \}_{q \neq p},
       \Theta).
       \end{align*}
Using Dirichlet--multinomial conjugacy, the probability vectors $\{\{
\boldsymbol{\phi}_n^{(p)}\}_{n=1}^{N^{(p)}(T)} \}_{p=1}^P$ can be
integrated out:
\begin{align*}
    &P(\mathcal{W} \g \Theta, \mathcal{T}, \mathcal{D}) = \notag \\
    &\quad\quad
\prod_{p=1}^P \prod_{n=1}^{N^{(p)}}
      \prod_{l=1}^{L^{(p)}_n} \frac{\sum_{l'=1}^{l-1}
        \mathbf{1}(w_{l',n}^{(p)} = w_{l,n}^{(p)}) +
        \alpha^{(p)}B^{(p)}_{w^{(p)}_{l,n},n}}{l - 1 + \alpha^{(p)}}. \label{eqs:bec-likelihood}
\end{align*}

To complete the specification of $P(\Theta)$, we place gamma
priors over the remaining parameters $\{\alpha^{(p)}, \boldsymbol{\beta}^{(p)}, \{ \rho^{(qp)} \}_{q \neq p}, \tau_L^{(p)}\}_{p=1}^{P}$.
The resultant posterior distribution $P(\Theta \g \mathcal{W},
\mathcal{T}, \mathcal{D})$ is analytically intractable; however,
posterior samples of $\{ \alpha^{(p)}, \boldsymbol{\beta}^{(p)}, \{ \rho^{(qp)} \}_{q \neq p},
\tau_L^{(p)} \}_{p=1}^P$ can be obtained using a collapsed
slice-within-Gibbs sampling algorithm~\citep{Neal2003}. Additional
details, including pseudocode, are given in the supplementary material.

\vspace{-0.2em}
\section{RELATED WORK}

Several recent probabilistic models use point processes as a
foundation for inferring influence and other social relationships from
temporal dynamics ~\citep{SimmaJordan2010, BluHelBec2012a,
  perry2013point, iwata2013discovering, dubois2013stochastic,
  zhou2013learning, linderman2014discovering}. Hawkes processes play
a central role in some of these models. Most relevant to this paper is the
work of~\citet{linderman2014discovering}, who used Hawkes processes to
study gang-related homicide in Chicago. Although temporal dynamics can
reveal some social relationships, others may be more readily evidenced
by also modeling interaction content. In this
vein,~\citet{danescu2012echoes} analyzed discussions among Wikipedians
and arguments before the US Supreme Court to uncover power
differences, while~\citet{gerrish10language-based} took a language-based approach to
measuring scholarly impact, identifying influential documents by
analyzing changes to thematic content over time.
These models differ significantly from ours, and have not been used
in a comparative analysis of different approaches to characterizing influence.
This paper compares approaches, demonstrating
that influence networks inferred from linguistic accommodation can be
more substantively meaningful than those inferred from
turn-taking. We also move beyond the work of
Danescu-Niculescu-Mizil et al. and Gerrish and Blei by defining a
generative, dynamic Bayesian language model that captures the mutual
excitation of words in social interactions.

\vspace{-0.2em}
\section{EXPERIMENTS}
\label{sec:experiments}

In this section, we showcase the Bayesian Echo Chamber's ability
to model transcripts of oral arguments heard by the US Supreme Court,
the transcript of the 1957 movie ``12 Angry Men,'' and meeting
transcripts from the Federal Reserve's Federal Open Market
Committee. We compare our model with competing approaches using the probability of held-out data,
and demonstrate that
our model can recover meaningful influence patterns from these data sets. We
also compare the linguistic accommodation-based influence networks inferred
 by our model to the turn-taking-based networks
inferred by Blundell et al.'s model.
Finally, we combine our model with Blundell et
al.'s in order to jointly model turn-taking and linguistic
accommodation. We also investigate tying the models' latent
influence parameters to determine whether a single global
notion of influence can be discovered.

The US Supreme Court consists of a chief justice and eight
associate justices. Each oral argument heard by the Court therefore
involves up to nine justices (some may recuse themselves) plus
attorneys representing the petitioner and the respondent. The format
of each argument is formulaic: the attorneys for each party have 30
minutes to present their argument, with those representing the
petitioner speaking first. Justices routinely interrupt the attorneys'
presentations to make comments or ask questions of the attorneys. Sometimes additional
attorneys, known as ``amicae curae,'' also present arguments in
support of either the petitioner or the respondent.
We used
the time-stamped
transcripts\footnote{\url{http://talkbank.org/data/Meeting/SCOTUS/}}
from three controversial Supreme Court
cases~\citep{macwhinney2007talkbank}: Lawrence and Garner v. Texas,
District of Columbia v. Heller, and Citizens United v. Federal
Election Commission (re-argument).

``12 Angry Men'' is a movie about a jury's deliberations regarding the
guilt or acquittal of a defendant. Unlike Supreme Court arguments, the
dialog is informal and intended to seem natural. The movie is unique in its limited cast of 12
people
and in the fact that it is set almost entirely in one
room. These qualities, combined with the fact that the movie
explicitly focuses on discussion-based consensus building in a group
setting, make its time-stamped transcript an ideal data set for
exploring the strengths of our model. We generated an appropriate
transcript from the movie subtitles by hand-labeling the person who made
each utterance.

The Federal Reserve's Federal Open Market
Committee oversees
the US's open market operations and sets the US national monetary
policy. The Committee consists of 12 voting members: the seven members
of the Federal Reserve Board and five of the 12 Federal Reserve Bank
presidents. By law, the FOMC must meet at least four times a year,
though it typically meets every five to eight weeks. At each meeting,
the Committee votes on the policy (tightening, neutrality, or easing)
to be carried out until the next meeting. Meeting transcripts are
embargoed for five years; as a result, transcripts from the meetings
surrounding the 2007--2008 financial crisis have only recently been
released. We used transcripts from 32 meetings ranging from March 27,
2006 to December 15, 2008,
inclusive.\footnote{\url{http://poliinformatics.org/data/}}

For all data sets, we concatenated consecutive utterances by the same
person, discarded contributions from people with fewer than ten
(post-concatenation) utterances, and rescaled all time stamps to the interval $(0, T=100]$. For each data set, we also restricted the
vocabulary to the $V=600$ most frequent stemmed word types. We did not
remove stop words, since they can carry important information about
influence relationships~\citep{danescu2012echoes}. The salient
characteristics of each data set, after preprocessing, are provided in
the supplementary material.

\vspace{-0.2em}
\subsection{Parameter Recovery}

\begin{figure}[!t]
  \centering
  \includegraphics[width=1.0\columnwidth]{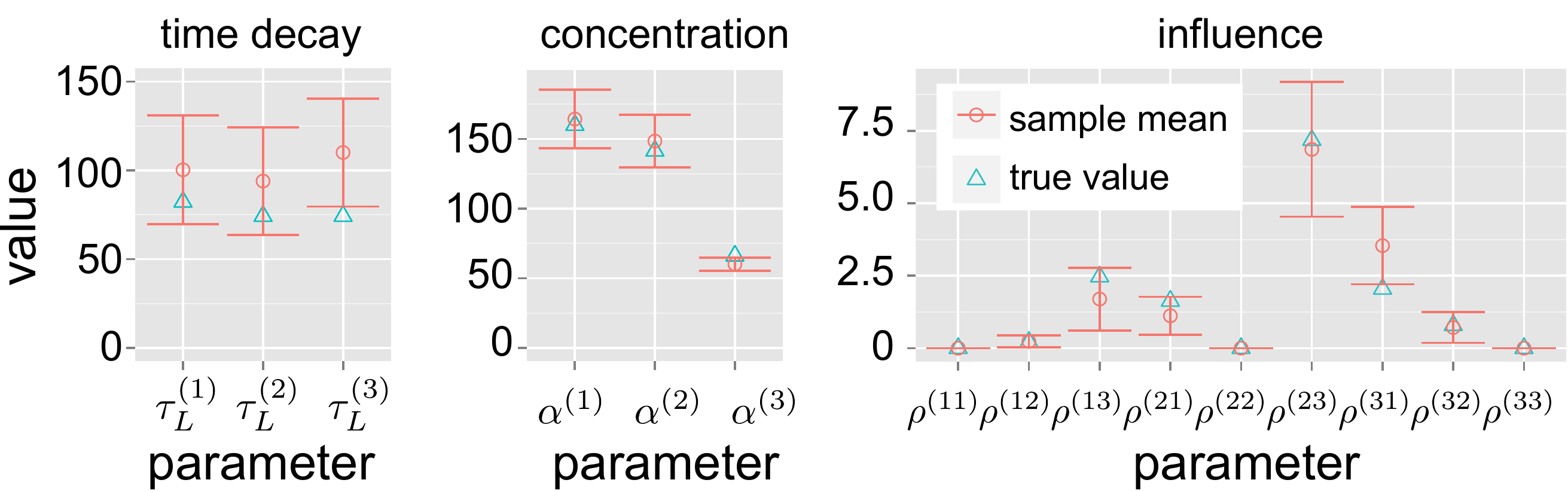}
  \caption{Parameter Recovery Results.}
  \label{fig:synthetic}
  \end{figure}

In this section, we present the results of a parameter recovery
experiment, conducted in order to validate our inference algorithm. We
used the generative process described in section~\ref{sec:model} to
generate 300 utterances made by $P=3$ people. In total, these
utterances contain 15,070 tokens spanning $V=20$ types. We drew the
length of each utterance, i.e., $L^{(p)}_n$, from a Poisson
distribution with a mean of 50. We generated the start times and
durations by assuming a round-robin approach to turn-taking and
setting the duration of each utterance to a value proportional to its
length in tokens. Figure~\ref{fig:synthetic} shows the true and
inferred parameter values, depicted using blue triangles and red
circles, respectively. The inferred parameter values were obtained
by averaging 3,000 samples from the posterior distribution. The error
bars indicate one standard deviation. Our proposed inference algorithm
does well at accurately recovering the true parameter values.

\vspace{-0.2em}
\subsection{Probability of Held-Out Data}
\label{sec:predictive-prob}

The predictive probability of held-out data, sometimes expressed as perplexity, is a standard metric for evaluating
statistical language models---the higher the probability, the better
the model. We compared predictive probabilities obtained
using the Bayesian Echo Chamber and several real-world data sets to
those obtained using two comparable language
models.

To compute the predictive probability of held-out data, we divided
each data set into a training set $\{ \mathcal{W}_{\textrm{train}},
\mathcal{T}_{\textrm{train}}, \mathcal{D}_{\textrm{train}}\}$ and a
held-out or test set $\{ \mathcal{W}_{\textrm{test}},
\mathcal{T}_{\textrm{test}}, \mathcal{D}_{\textrm{test}}\}$. We formed
each training set by selecting those utterances that occurred before
some time $t*$ where $t*$ was chosen to yield either a 90\%--10\% or
80\%--20\% training--testing split, i.e.,
$\mathcal{W}_{\textrm{train}} = \{ \boldsymbol{w}_n^{(p)} :
{t'}_n^{(p)} \leq t* \}_{p=1}^P$ and $\mathcal{W}_{\textrm{test}} = \{
\boldsymbol{w}_m^{(p)} : {t}_m^{(p)} > t* \}_{p=1}^P$.  The predictive
probability of held-out data is then $P(\mathcal{W}_{\textrm{test}} \g
\mathcal{T}_{\textrm{test}}, \mathcal{D}_{\textrm{test}},
\mathcal{W}_{\textrm{train}}, \mathcal{T}_{\textrm{train}},
\mathcal{D}_{\textrm{train}})$. Although this probability is
analytically intractable, its logarithm can be approximated via the
lower bound $\frac{1}{S} \sum_{s=1}^S
\log{P(\mathcal{W}_{\textrm{test}} \g \mathcal{T}_{\textrm{test}},
  \mathcal{D}_{\textrm{test}}, \Theta^{(s)})}$ where $\Theta^{(s)}$
denotes a set of sampled parameter values drawn from the posterior
distribution $P(\Theta \g \mathcal{W}_{\textrm{train}},
\mathcal{T}_{\textrm{train}}, \mathcal{D}_{\textrm{train}})$.

Approximate log probabilities obtained using the Bayesian Echo Chamber (with $S=3000$
samples after 1000 burn-in sampling iterations), a unigram language
model, and Blei and Lafferty's dynamic topic
model~\citeyearpar{blei06dynamic} are provided in
table~\ref{tab:log_prob}. Log probabilities for additional data sets
are provided in the supplementary material. The unigram language model
is equivalent to setting all influence parameters in our model to
zero. In all experiments involving the dynamic topic model, each data
set was sliced into $K=10$ or $K=5$ equally-sized time slices
(depending on the training--testing split), with the last slice taken
to be the test set and utterances treated as documents. Each log
probability reported for the dynamic topic model is the highest value
obtained using either 5, 10, or 20 topics. Since inference for the
dynamic topic model was performed using a variational inference
algorithm,\footnote{Inference code obtained from
  \url{http://www.cs.princeton.edu/~blei/topicmodeling.html}} its log
probabilities are also lower bounds and standard deviations are not
available. For all data sets, the Bayesian Echo Chamber out-performed
both the unigram language model and the dynamic topic model.

\label{sec:logprob}
\begin{table*}[htb]
  \caption{Predictive Log Probabilities of Held-Out Data.}
  \label{tab:log_prob}
  \scriptsize
  \centering
\begin{tabular}{@{}lr@{$\pm$}lr@{$\pm$}lrr@{$\pm$}lr@{$\pm$}lr@{}}
\toprule
& \multicolumn{5}{c}{10\% Test Set}                                         & \multicolumn{5}{c}{20\% Test Set}                                   \\  \cmidrule(l){2-11}
Data Set                               & \multicolumn{2}{c}{Our Model}  & \multicolumn{2}{c}{Unigram} & DTM~~           & \multicolumn{2}{c}{Our Model}   & \multicolumn{2}{c}{Unigram} & DTM~~     \\
\midrule
Synthetic                      & \textbf{-4292.97}          & 0.02 & -4297.92       & 0.04       & -4364.81           & \textbf{-8702.92} & 0.04 & -8717.77       & 0.08      & -8948.07  \\
DC v.~Heller                   & \textbf{-7383.45} & 0.12 & -7794.25       & 0.21       & -7533.58         & \textbf{-12404.21} & 0.15 & -13126.73       & 0.26      & -12744.73 \\
L\&G v.~Texas                  & \textbf{-6663.33} & 0.12 & -6937.66       & 0.18       & -6759.06         & \textbf{-10248.80} & 0.21 & -10791.25       & 0.23      & -10459.87 \\
Citizens United v.~FEC         & \textbf{-5770.12} & 0.14 & -6120.67       & 0.18       & -5851.224        & \textbf{-16370.7}  & 0.95 & -17157.21       & 0.40      & -16400.46 \\
``12 Angry Men''                   & \textbf{-4667.47} & 0.24 & -4920.21       & 0.14       & -4691.11         & \textbf{-8722.97}  & 0.27 & -9222.99        & 0.25      & -8787.35  \\ \bottomrule
\end{tabular}
  \end{table*}

\vspace{-0.2em}
\subsection{Influence Recovery}

In this section, we demonstrate that the Bayesian Echo Chamber can
recover known influence patterns in Supreme Court arguments and in the
movie ``12 Angry Men.'' We also use these data sources to to compare
influence networks inferred by our model to those inferred by the
model described in section~\ref{sec:turn-taking}. All reported influence parameters were obtained by
averaging 3,000 posterior samples; posterior standard deviations are
provided in the supplementary material.

\begin{figure*}[!htb]
\centering
  \subfloat[]{
    \includegraphics[width=0.33\textwidth]{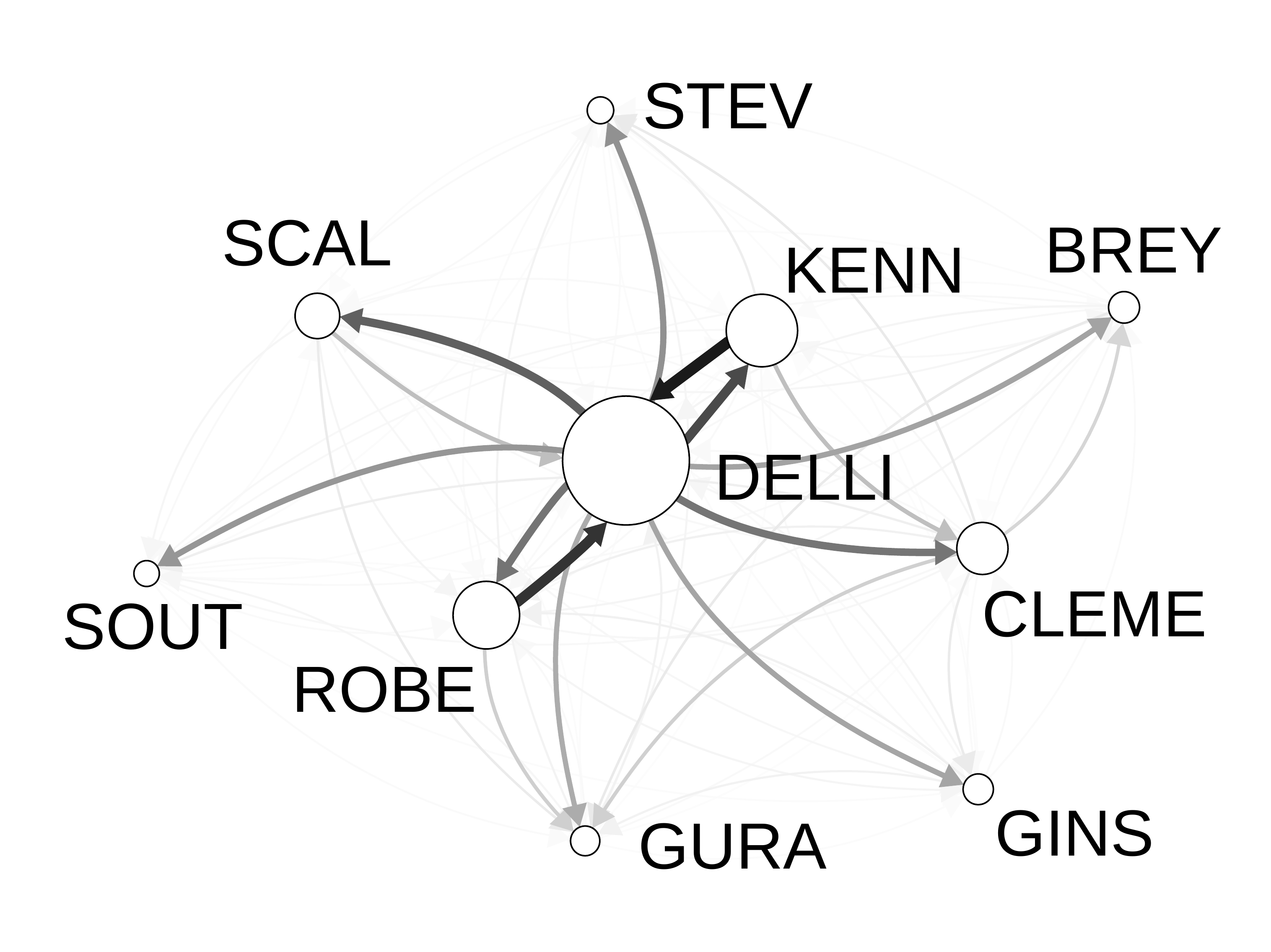}
    }
  \subfloat[]{
    \includegraphics[width=0.33\textwidth]{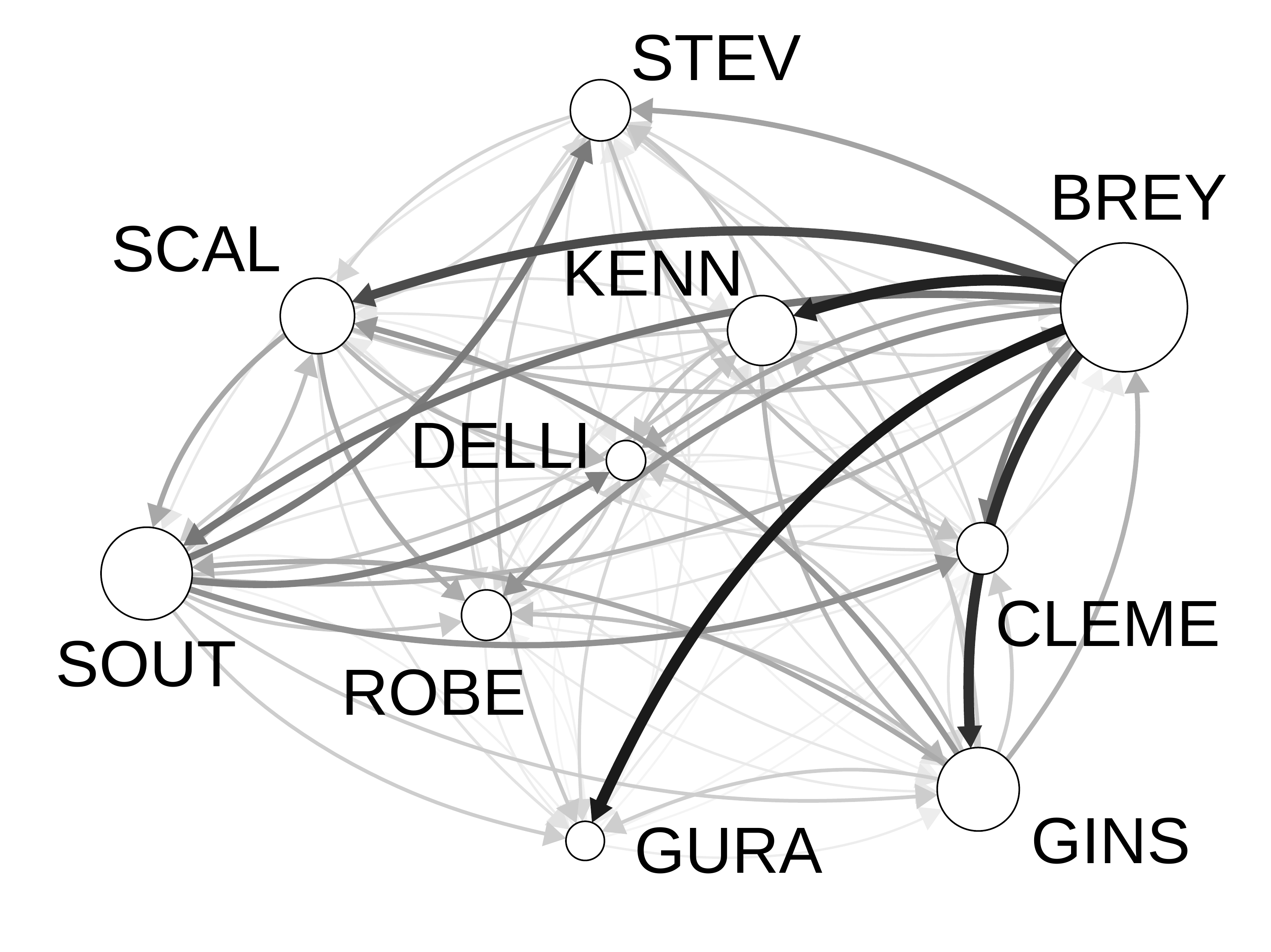}
    }
  \subfloat[]{
    \includegraphics[width=0.26\textwidth]{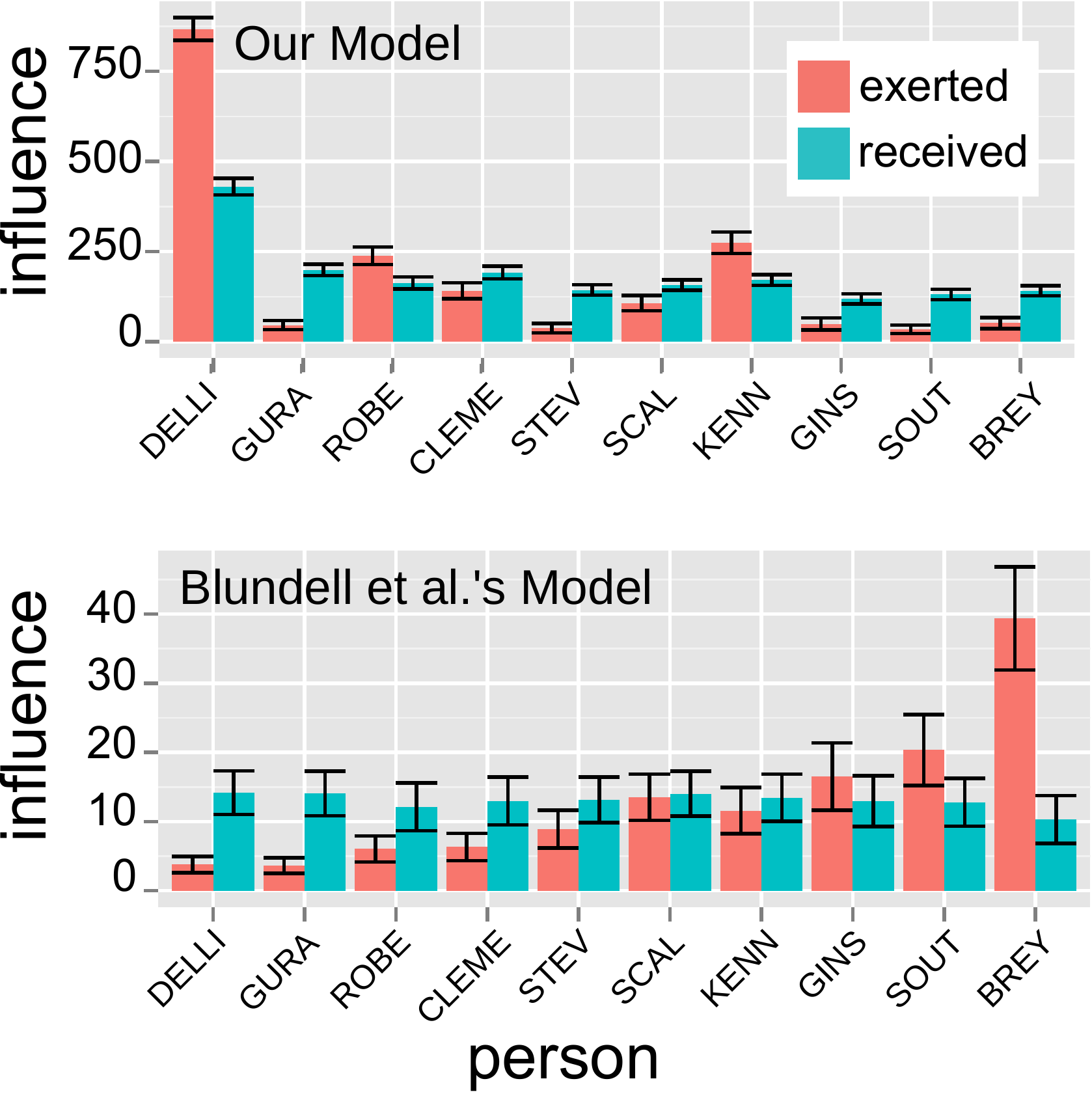}
    }
\caption{Influence Networks (Posterior Mean) Inferred from the DC v. Heller Case.
  (a) Network Inferred Using Our Model and (b) Inferred Using Blundell
  et al.'s Model.
(c) Total Influence Exerted/Received by Each Person.}
\label{fig:DCvsHeller}
\end{figure*}

\vspace{-0.2em}
\subsubsection{US Supreme Court}

As described previously, Supreme Court arguments are extremely
formulaic: The attorneys representing the petitioner present their
argument first, speaking for a total of 30 minutes before the
respondent's attorneys are allowed to present their argument. Justices
routinely interrupt these presentations. We therefore anticipate that
influence networks inferred from linguistic accommodation patterns
will reveal significant influence exerted by the petitioner's
attorneys, simply because they speak first, establishing the language
used in the rest of the discussion. We also anticipate that influence
networks inferred from turn-taking behavior will reveal significant
influence exerted by the justices over the attorneys. This is because
the justices interrogate the attorneys' during their presentations.

As an illustrative example, we present results obtained from the
District of Columbia v. Heller case in
figure~\ref{fig:DCvsHeller}. (The other two cases, Lawrence and Garner
v. Texas and Citizens United v. Federal Election Commission exhibited
remarkably similar influence networks.)
The influence network\footnote{Plotted using \texttt{qgraph}~\citep{epskamp2012qgraph}.} inferred
using the Bayesian Echo chamber is shown in~\ref{fig:DCvsHeller}(a),
while the network inferred using Blundell et al.'s model is
shown in~\ref{fig:DCvsHeller}(b). To illustrate posterior uncertainty, networks drawn with different posterior quantiles are provided in the supplementary material.
The total influence exerted and
received by each participant are shown for each model in figure~\ref{fig:DCvsHeller}(c). The error bars represent the posterior standard deviation.
The justices present for this case
were Alito, Breyer, Ginsburg, Kennedy, Roberts, Scalia, Stevens,
Souter, and Thomas, while the attorneys were Dellinger (representing
the petitioner), Gura (representing the respondent), and Clement (as
amicae curae, supporting the petitioner). Ultimately, Alito, Kennedy,
Roberts, Scalia, and Thomas (the majority) sided with the respondent,
while Breyer, Ginsburg, and Stevens (the minority) sided with the
petitioner. Neither Alito or Thomas spoke ten or more utterances, so
they were not included in our analyses.

The influence network inferred using our model is very sparse. As
expected, Dellinger (who represented the petitioner and presented his
argument first) is shown as exerting the most influence. The justices
with the most influence are Kennedy and Roberts, both of whom
ultimately supported the respondent and thus interrogated Dellinger
much more the other justices.

The most striking pattern in the influence network inferred using
Blundell et al.'s model is that the three attorneys received much more
influence from the justices than vice versa. This pattern could
be seen as reflecting the status difference between justices and
attorneys or as reflecting the formulaic structure of the
Supreme Court: attorneys present arguments, while justices interrupt
to make comments or ask questions.

\vspace{-0.2em}
\subsubsection{``12 Angry Men''}

Unlike Supreme Court arguments, the dialog in ``12 Angry Men'' is
informal and intended to seem natural. Since the focus of the movie is
discussion-based consensus building in a group setting, we therefore
anticipate that the narrative of the movie will be
reflected in influence networks inferred from linguistic
accommodation patterns and from turn-taking behavior.

\begin{figure*}[!htb]
\centering
  \subfloat[]{
    \includegraphics[width=0.33\textwidth]{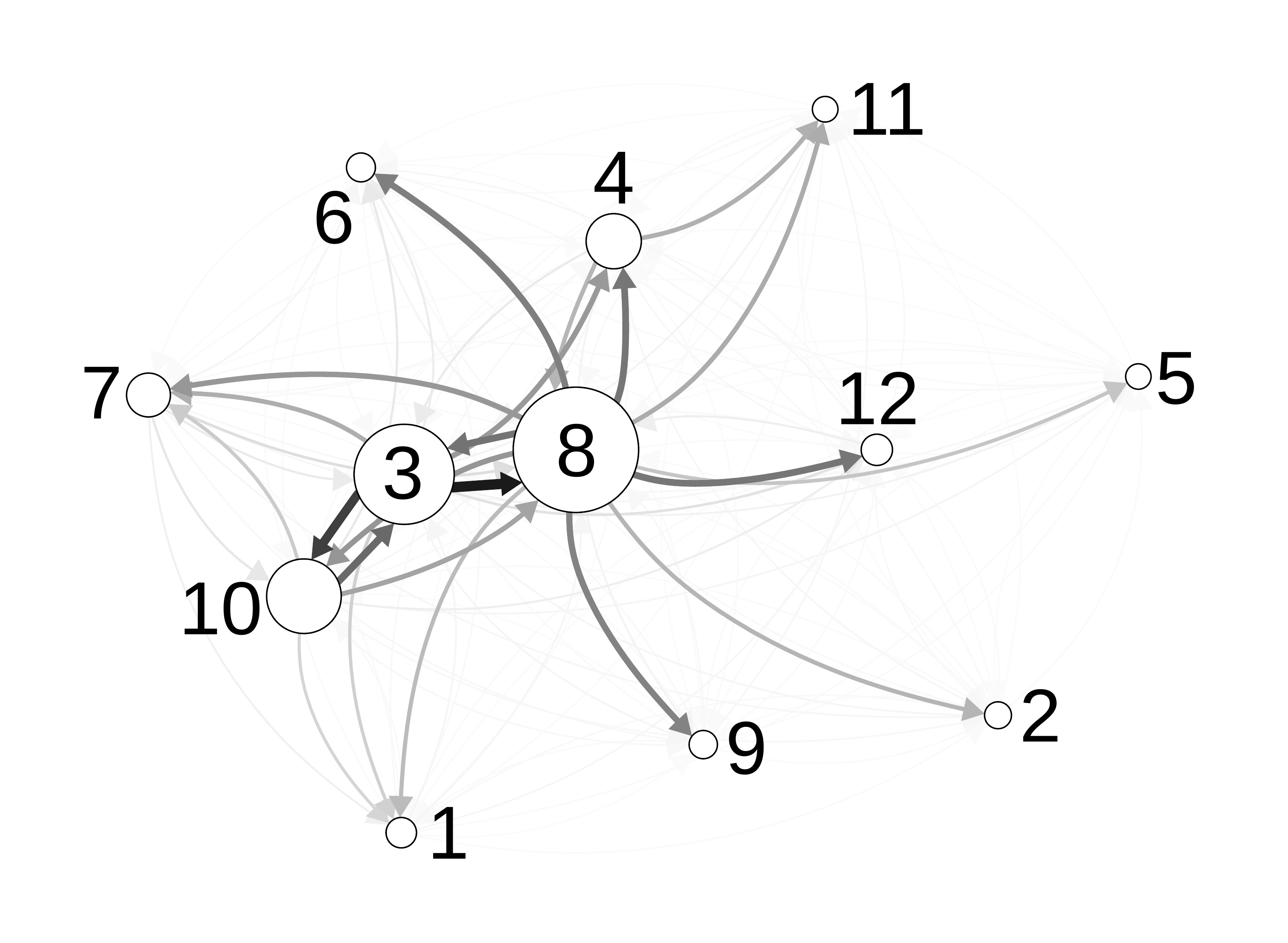}
    }
  \subfloat[]{
    \includegraphics[width=0.33\textwidth]{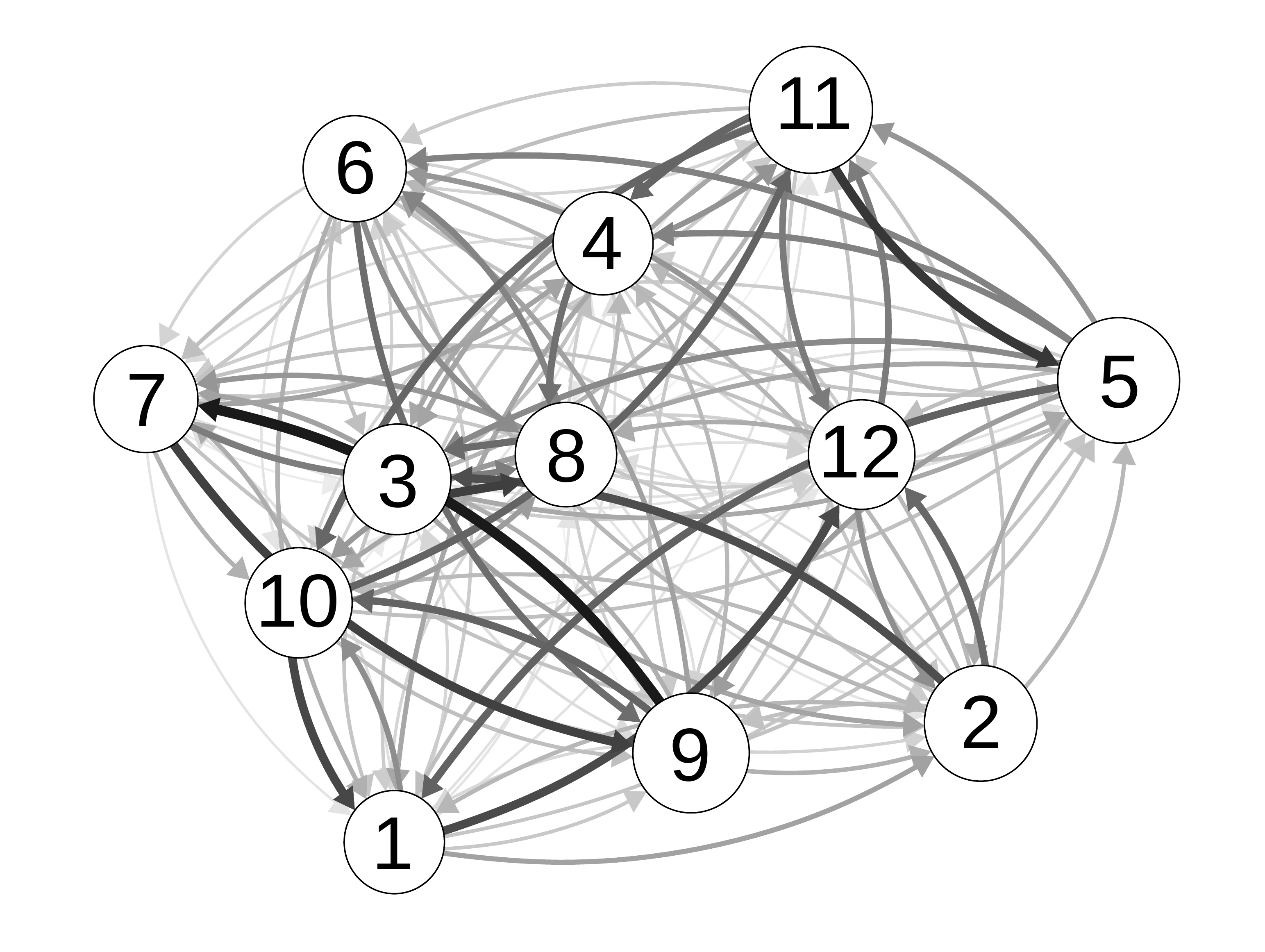}
    }
  \subfloat[]{
    \includegraphics[width=0.28\textwidth]{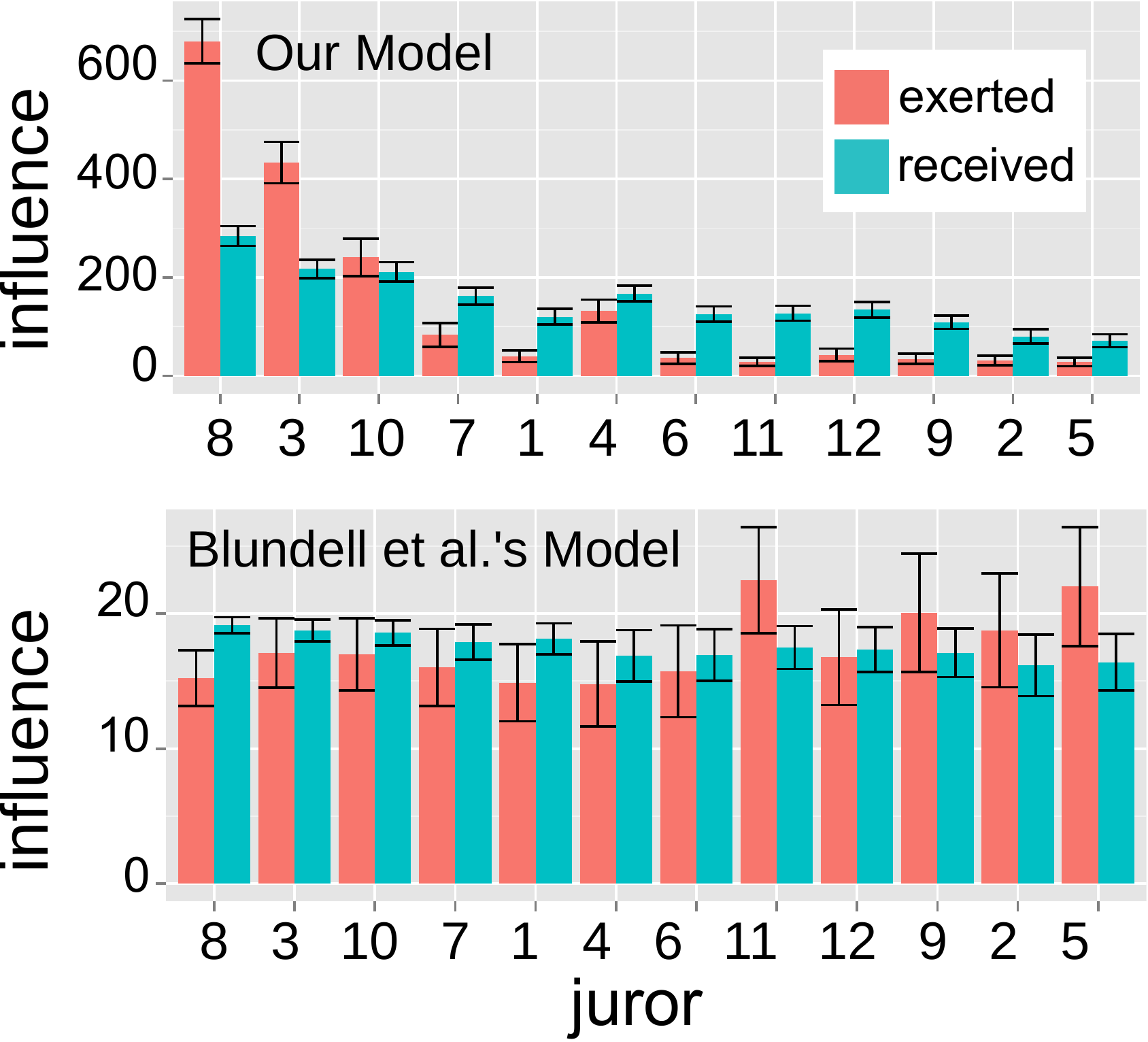}
    }
\caption{Influence Networks (Posterior Mean) Inferred from ``12 Angry Men.''
  (a) Network Inferred Using Our Model and (b) Inferred Using Blundell
  et al.'s Model.
(c) Total Influence Exerted/Received by Each Person.}
\label{fig:12AngryMen}
\end{figure*}

The influence network inferred using our model is shown in
figure~\ref{fig:12AngryMen}(a), while the total influence exerted and
received by each juror are shown in the top of figure
~\ref{fig:12AngryMen}(c). The most significant pattern is that three
individuals exert more influence over others than others do over them:
Juror 8, Juror 3, and, to a lesser extent, Juror 10. Juror 8 is the
protagonist of the movie, and initially casts the only ``not guilty''
vote. The other jurors ultimately change their votes to match
his. Juror 3, the antagonist, is the last to change his vote. It
therefore unsurprising that Juror 8, the first to vote ``not guilty'',
should dominate the discussion content. Similarly, Juror 3, the last
to change his ``guilty'' vote, is most invested in discussing
defendant's supposed guilt. Juror 10 is one
of the last three jurors, along with Jurors 3 and 4, to change his vote. However, unlike Juror 4
(who stands out marginally in figure~\ref{fig:12AngryMen}(a) and,
according to figure~\ref{fig:12AngryMen}(c), has less influence over
others than others do over him), Juror 10 is argumentative as he
changes his mind. Overall, the consistency of the inferred influence
network with the narrative of the movie confirms that the Bayesian
Echo Chamber can indeed uncover substantive influence relationships.

The influence network inferred using Blundell et al.'s model and the
total influence exerted and received by each juror are shown in
figure~\ref{fig:12AngryMen}(b) and the bottom of
\ref{fig:12AngryMen}(c), respectively. The four jurors who exert more
influence over others than others do over them (Juror 2, Juror 5,
Juror 9, and Juror 11) are the first four jurors to change their
votes. Jurors 5 and 11, who exert the most influence, are verbose,
while Jurors 2 and 9 are comparatively taciturn. Jurors 8 exerts
little influence because he must respond to questions and defend his
position as he tries to persuade the others to agree with him, much
like the attorneys in the Supreme Court.

\vspace{-0.2em}
\subsection{Exploratory Analysis of FOMC Meetings}

\begin{figure*}[!htb]
\centering
  \subfloat[]{
    \includegraphics[width=0.284\textwidth]{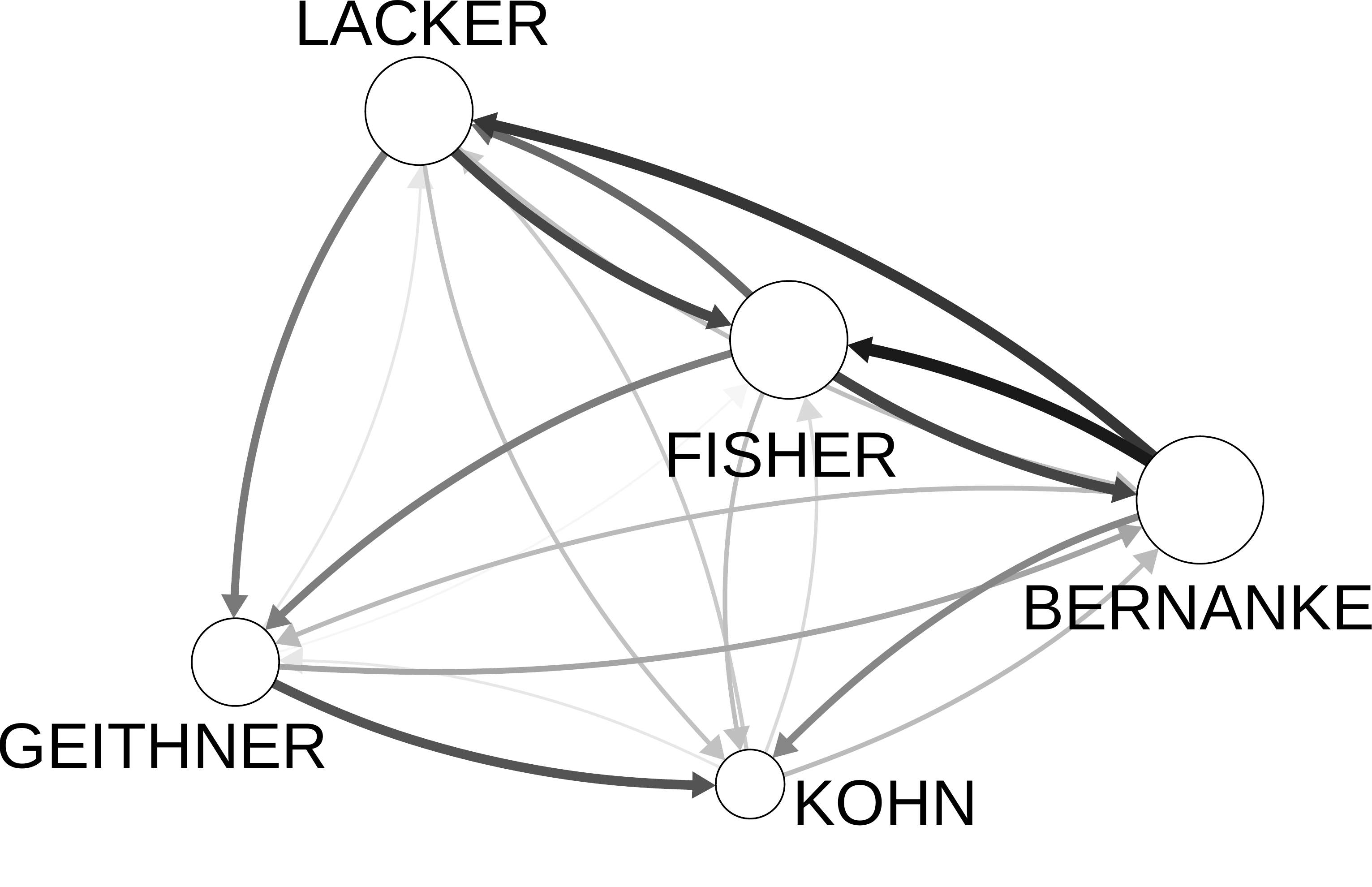}
    }
  \subfloat[]{
    \includegraphics[width=0.284\textwidth]{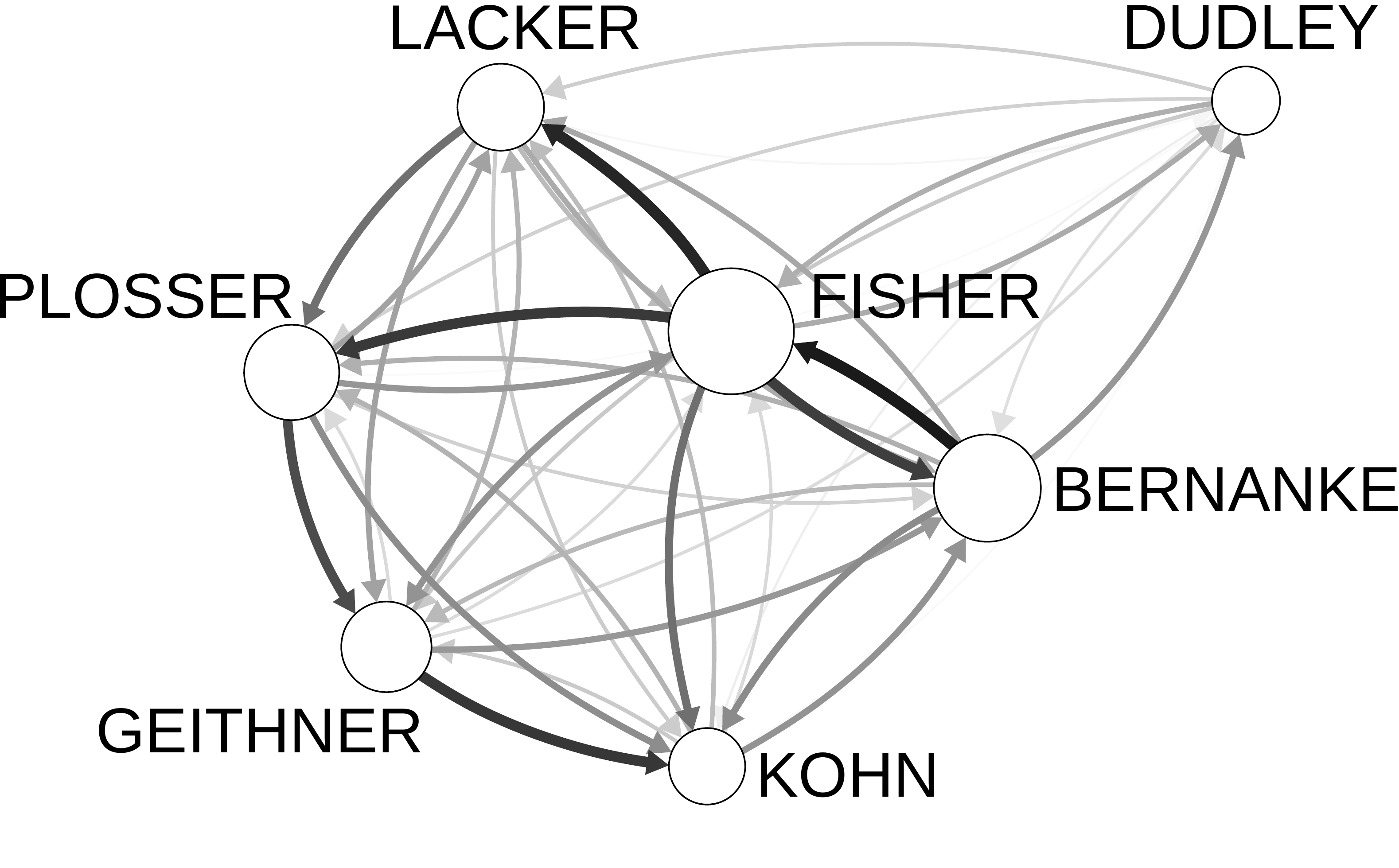}
    }
  \subfloat[]{
    \includegraphics[width=0.284\textwidth]{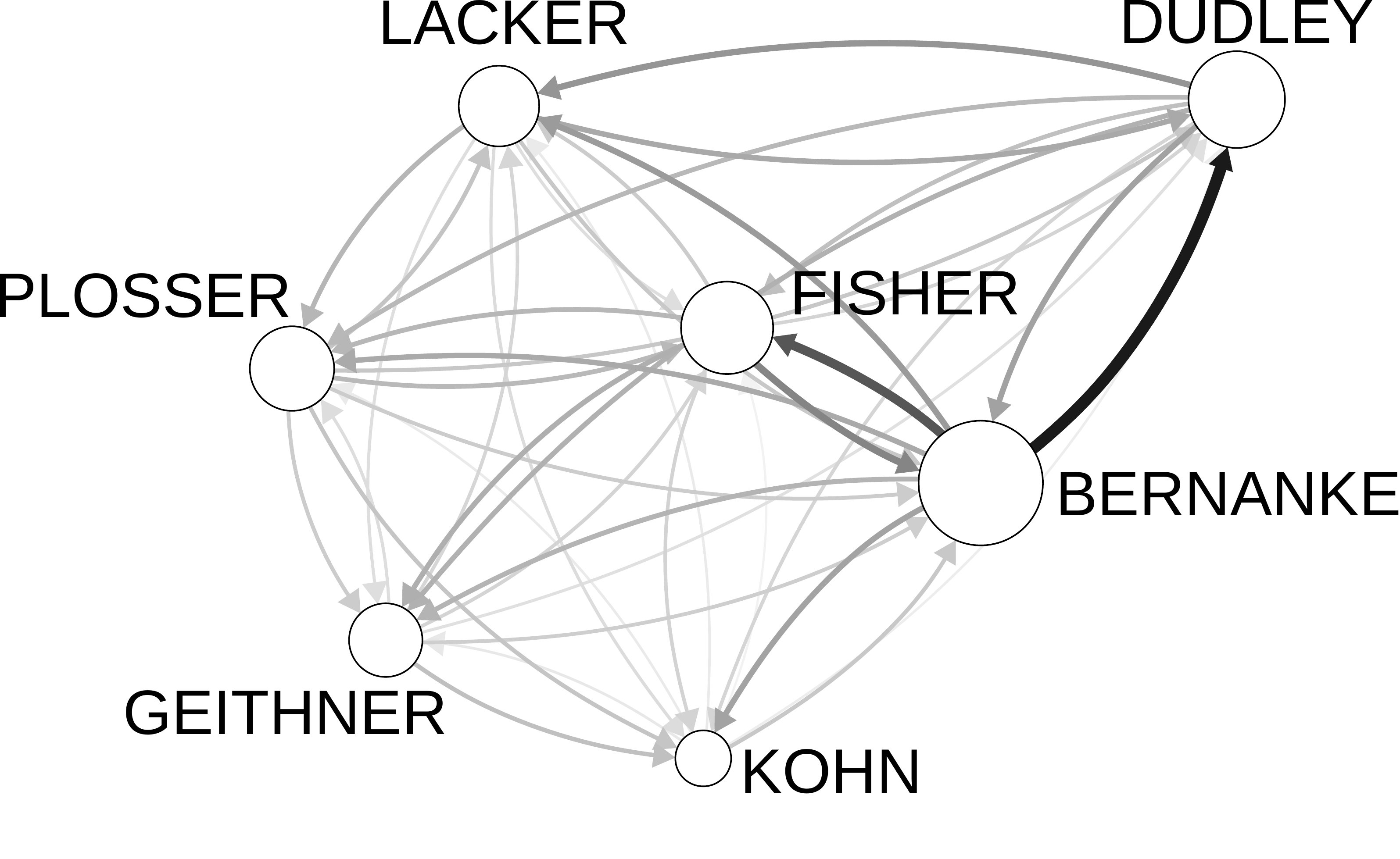}
    }
\caption{Influence Networks (Posterior Mean) Inferred from FOMC Meetings
  Using Our Model. (a)
  March 27, 2006--June 28, 2006. (b) August 8, 2006--August 7, 2007. (c) August 10, 2007--December 15, 2008.}
\label{fig:fomc-bernanke}
\end{figure*}

Finally, we performed an exploratory analysis of the relationships
inferred from transcripts of 32 Federal Open Market Committee meetings
surrounding the 2007--2008 financial crisis, ranging from March 27,
2006 to December 15, 2008, inclusive. Since utterance durations are
not available for these transcripts (also preventing the use of Blundell et al.'s model), we set the duration of each
utterance to a value proportional to its length in
tokens. We divided
the meetings into three subsets: March 27, 2006 through June 28, 2006;
August 8, 2006 through August 7, 2007; and August 10, 2007 through
December 15, 2008. The first subset corresponds to meetings with a
resultant policy of tightening; the second to meetings with a neutral
outcome; and the third to meetings that resulted in easing. These
meetings were all chaired by Bernanke.

Figure~\ref{fig:fomc-bernanke} depicts the influence network for each
subset (aggregated by averaging over the meetings in that subset) inferred using our model. In the first network, corresponding to
pre-crash meetings from March 27, 2006 through June 28, 2006,
Bernanke, Fisher, and Lacker play the biggest roles with Bernanke, the
chair, exerting the most influence over others. Given his role as
chair, Bernanke's involvement is arguably unsurprising, but Fisher and
Lacker's roles are notable. Unlike Bernanke, Fisher and Lacker are
both ``hawks'' and thus generally in favor of tightening monetary
policy; the meetings in this subset all resulted in an outcome of
tightening. In the second network, corresponding to pre-crash meetings
from August 8, 2006 through August 7, 2007, Bernanke, Fisher, and
Lacker all continue to play significant roles, but the network is much
less sparse, with both hawks and ``doves'' (those generally in favor of
easing monetary policy) exerting influence over others. In contrast to
the meetings in the previous subset, these meetings resulted in
neutrality---i.e., neither tightening or easing. Finally, in the third
network, corresponding to post-crash meetings from August 10, 2007
through December 15, 2008, there are fewer strong influence
relationships. Bernanke (the chair and a dove) still plays a major
role, while Fisher and Lacker's roles are significantly
diminished. Instead, Dudley, also a dove and a close ally of Bernanke,
plays a much greater role, especially in his relationship with
Bernanke. These meetings all resulted in monetary policy easing, a
strategy generally favored by doves and opposed by hawks.

There has been little work in political science,
economics, or computer science on analyzing these
meeting transcripts. As a result, the inferred networks not only showcase our
model's ability to discover latent influence relationships from
linguistic accommodation, but also constitute a research
contribution of substantive interest to political
scientists, economists, and other social scientists studying the financial crisis.

\vspace{-0.2em}
\subsection{Model Combination}

Since influence can be inferred from both turn-taking behavior and
linguistic accommodation, we explored the possibility of combining the Bayesian Echo Chamber and Blundell et al.'s model
to form a ``supermodel'' with a single set of shared influence
parameters. The simplest way to share these parameters is to tie
them together as
$\rho^{(qp)} = r \nu^{(qp)}$, where $r$ is a scaling factor and
$\rho^{(qp)}$ and $\nu^{(qp)}$ correspond to the influence from person $q$ to person $p$
in our model and Blundell et al.'s model, respectively.
Tying the influence parameters in
this way provides the model with the capacity to capture a global
notion of influence that is based upon both turn-taking and linguistic
accommodation.

This tied model,
whose likelihood is the product of the Bayesian Echo Chamber's likelihood and that of
Blundell et al.'s model but with shared influence parameters,
assigned lower probabilities to held-out data than the fully
factorized model (i.e., separate influence parameters). Log
probabilities, obtained using a 90\%--10\%
training--testing split and a vocabulary of $V=300$ word types in order
to reduce computation time, are provided in the supplementary material.


Interestingly, the networks inferred by the model with tied parameters
are
extremely similar to those inferred using the Bayesian Echo
Chamber. These results suggest that linguistic accommodation reflects
a more informative notion of influence that that evidenced via
turn-taking. We expect that investigating other ways of
combining turn-taking-based models with ours will be a promising
direction for future exploration.


\vspace{-0.2em}
\section{DISCUSSION}

The Bayesian Echo Chamber is a new generative model for discovering
latent influence networks via linguistic accommodation patterns. We
demonstrated that our
model can recover known influence patterns in synthetic data,
arguments heard by the US Supreme Court, and in the movie
``12 Angry Men.'' We compared influence networks inferred using our
model to those inferred using a variant of Blundell et al.'s turn-taking-based model
and showed that by modeling linguistic accommodation
patterns, our model infers different, and often more meaningful, influence
networks. Finally, we showcased our model's potential as an
exploratory analysis tool for social scientists by inferring latent
influence relationships between members of the
Federal Reserve's Federal Open Market Committee.

Promising avenues for future work include (1) modeling
linguistic accommodation separately for function and content words and
(2) explicitly modeling the dynamic evolution of influence networks over time.

\subsubsection*{Acknowledgements}

Thanks to Juston Moore and Aaron Schein for their work on early stages
of this project, and to Aaron for the ``Bayesian Echo Chamber''
model name. This work was supported in part by the Center for Intelligent
Information Retrieval, in part by NSF grant \#IIS-1320219, and in part
by NSF grant \#SBE-0965436. Any opinions, findings and conclusions or
recommendations expressed in this material are the authors' and do not
necessarily reflect those of the sponsor.

\bibliographystyle{apalike}
\bibliography{refs}

\thispagestyle{empty}
\setcounter{section}{0}

\twocolumn[

\aistatstitle{Supplementary Material for ``The Bayesian Echo Chamber''}

\aistatsauthor{Fangjian Guo \And Charles Blundell \And Hanna Wallach
  \And Katherine Heller}

\aistatsaddress{ Duke University\\Durham, NC,
  USA\\ {\texttt{guo@cs.duke.edu}} \And Gatsby Unit, UCL\\London, UK\\ {\texttt{c.blundell@gatsby.ucl.ac.uk~~}} \And
  Microsoft Research\\New York, NY, USA\\ {\texttt{~~wallach@microsoft.com}} \And Duke
  University\\Durham, NC, USA\\ {\texttt{kheller@stat.duke.edu}}} ]

\section{INFLUENCE VIA TURN-TAKING}

In this section, we provide appropriate priors and details of an
inference algorithm for the variant of Blundell et al.'s
model~\citeyearpar{BluHelBec2012a} described in section 2 of the
paper. For real-world group discussions, the utterance start times
$\mathcal{T} = \{ \mathcal{T}^{(p)} \}_{p=1}^P$ and durations
$\mathcal{D} = \{ \{ \Delta t_n^{(p)}\}_{n=1}^{N^{(p)}(T)}\}_{p=1}^P$
are observed, while parameters $\Theta = \{ \lambda_0^{(p)}, \{
\nu^{(qp)}\}_{q \neq p}, \tau_T^{(p)} \}_{p=1}^P$ are unobserved;
however, information about the values of these parameters can be
quantified via their posterior distribution given $\mathcal{T}$ and
$\mathcal{D}$, obtained via Bayes' theorem, i.e.,
\begin{equation}
  P(\Theta \g \mathcal{T}, \mathcal{D}) \propto
P(\mathcal{T} \g \Theta, \mathcal{D})\,P(\Theta).
\end{equation}

The likelihood term has the form
\begin{align}
&P(\mathcal{T} \g \Theta, \mathcal{D}) =\notag\\
  &\quad \prod_{p=1}^P \left(
\exp{\left( - \Lambda^{(p)}(T) \right)}
\prod_{n=1}^{N^{(p)}(T)} \lambda^{(p)}(t_n^{(p)})
\right),
\label{eqn:hawkes_likelihood}
 \end{align}
where $\Lambda^{(p)}(T) = \int_0^T \lambda^{(p)}(t)\,\textrm{d}t$ is
the expected total number of utterances made over the entire
observation interval from $0$ to $T$~\citep{daley88introduction}.

  Like Blundell et al.,  we place an improper prior over
  $\lambda_0^{(p)}>0$.  We also use priors to ensure that the
  multivariate Hawkes process is stationary. Specifically, we employ
  the stationarity condition of \citet{bremaud96stability}. If
  $\boldsymbol{M}$ is a $P \times P$ matrix given by
\begin{equation}
M^{(qp)} = \int_u^{\infty} \left|\, g^{(qp)}(t, u) \right| \,\textrm{d}t= \nu^{(qp)} \tau^{(p)}_T,
\end{equation}
then this condition requires the spectral radius of
$\boldsymbol{M}$ to be strictly less than one. This condition is not
straightforward to enforce with tractable constraints; however, since
the spectral radius of $\boldsymbol{M}$ is upper-bounded by any matrix
norm, the condition may be enforced by requiring that $\|
\boldsymbol{M} \| < 1$ for any norm $\| \cdot \|$. We
use the maximum absolute column sum norm:
\begin{align}
\| \boldsymbol{M} \|_{1 \rightarrow 1}&= \max_{\|x\|_1=1}\|\boldsymbol{M}x\|_{1}\\ &= \max_{p=1,\cdots,P} \tau_{T}^{(p)} \sum_{q\neq p}\nu^{(qp)}.\label{eqn:condition}
\end{align}
Rewriting this expression implies an improper joint prior over
$\{\tau_{T}^{(p)}\}_{p=1}^P$ and $\{\{\nu^{(qp)}\}_{q \neq
  p}\}_{p=1}^P$ in which
\begin{align}
& 0<\tau_{T}^{(p)}<\frac{1}{\sum_{q \neq
      p} \nu^{(qp)}} \textrm{ and } \\
&\quad 0<\nu^{(qp)}<\frac{1}{\tau_{T}^{(p)} - \sum_{r \neq q, r \neq p}} \nu^{(rp)}.
\end{align}

Although the resultant posterior distribution $P(\Theta \g
\mathcal{T}, \mathcal{D})$ is analytically intractable, posterior
samples can be drawn using either the conditional intensity function
approach or the cluster process approach described
by~\citet{rasmussen2013bayesian}. Like Blundell et al., we take the
former approach and use a slice-within-Gibbs
algorithm~\citep{Neal2003} that sequentially samples each parameter
from its conditional posterior.

This slice-within-Gibbs algorithm requires frequent evaluation of the
likelihood in equation~\ref{eqn:hawkes_likelihood}; however, the
computational cost can be reduced by noting that the product over rate
functions can be efficiently computed using the following recurrence
relation:
\begin{align*}
  &\lambda^{(p)}(t_n^{(p)}) = \notag \\
  &\quad \lambda_0^{(p)} + \left( \lambda^{(p)}(t_{n-1}^{(p)}) -
  \lambda_0^{(p)}\right) \exp{\left( -\frac{t_n^{(p)} -
      t_{n-1}^{(p)}}{\tau_T^{(p)}} \right)} + {} \notag\\
  &\quad \sum_{q \neq p} \sum_{m: t_{n-1}^{(p)} \leq {t'}_m^{(q)} <
    t_n^{(p)}} \nu^{(qp)} \exp{\left( - \frac{t_n^{(p)} - {t'}_m^{(q)}}{\tau_T^{(p)}} \right)}
  \end{align*}
for $n=2, 3, \ldots, N^{(p)}(T)$. The initial term is
\begin{align*}
&\lambda^{(p)}(t_1^{(p)}) = \notag\\
  &\quad \lambda_0^{(p)} + \sum_{q\neq
  p}\sum_{m:{t'}_{m}^{(q)} < t_{1}^{(p)} }\nu^{(qp)} \exp{\left( -\frac{t^{(p)}_{1}-{t'}_m^{(q)}}{\tau_T^{(p)}}\right)}.
\end{align*}

\section{INFLUENCE VIA LINGUISTIC ACCOMMODATION}

In this section, we provide a directed graphical model, appropriate
priors, and details of an inference algorithm for our model, the
Bayesian Echo Chamber.

\begin{figure}[!thb]
\centering\includegraphics[width=0.9\columnwidth]{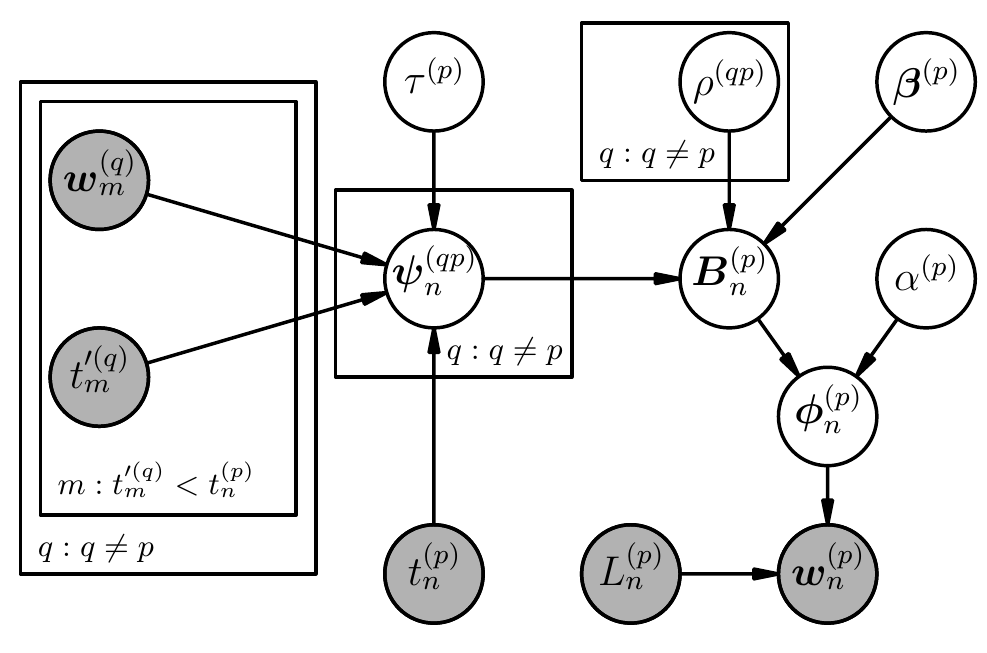}
\caption{Directed Graphical Model Depicting the Structure of $P(\boldsymbol{w}_n^{(p)} \g
         \{\{\boldsymbol{w}_m^{(q)}\}_{m : {t'}_m^{(q)}
                       < t^{(p)}_n} \}_{q \neq p}, \Theta)$.}
\label{fig:pgm}
\end{figure}

The likelihood term implied by our model is
\begin{align*}
    &P(\mathcal{W} \g \Theta, \mathcal{T}, \mathcal{D}) = \\
    &\quad\prod_{p=1}^P \prod_{n=1}^{N^{(p)}(T)}
  P(\boldsymbol{w}_n^{(p)} \g
         \{\{\boldsymbol{w}_m^{(q)} \}_{m: {t'}_m^{(q)}
                < t^{(p)}_n} \}_{q \neq p},
                \Theta).
                       \end{align*}
A directed graphical model depicting the structure of
$P(\boldsymbol{w}_n^{(p)} \g \{\{\boldsymbol{w}_m^{(q)}\}_{m :
{t'}_m^{(q)} < t^{(p)}_n} \}_{q \neq p}, \Theta)$ is in
figure~\ref{fig:pgm}.

We place a gamma prior over $\rho^{(qp)}$, with a shape parameter
chosen to encourage shrinkage towards zero. Due to the additive nature
of $\boldsymbol{B}_n^{(p)}$, the value of $\beta_v^{(p)}$ should be
comparable in magnitude to $\sum_{q \neq p} \rho^{(qp)}
\psi_{v,n}^{(qp)}$. We therefore place a gamma prior over each
$\beta_v^{(p)}$, with shape and scale parameters chosen to yield this
property for real-world data sets. We also place broad gamma priors over
$\alpha^{(p)}$ and $\tau_L^{(p)}$. In practice, inference is
insensitive to the specific values of the shape and scale parameters
of these priors, provided they are broad. For our experiments, we
used $\alpha^{(p)} \sim \textrm{Gamma}\,(10, 10)$, $\beta_v^{(p)} \sim
\textrm{Gamma}\,(10, 20)$, $\rho^{(qp)} \sim \textrm{Gamma}\,(1, 2)$, and $\tau^{(p)} \sim \textrm{Gamma}\,(10, 10)$.

Although the resultant posterior distribution $P(\Theta \g
\mathcal{W}, \mathcal{T}, \mathcal{D})$ is intractable, posterior
samples of $\{ \alpha^{(p)}, \boldsymbol{\beta}^{(p)}, \{ \rho^{(qp)}
\}_{q \neq p}, \tau_L^{(p)} \}_{p=1}^P$ can be drawn using a
collapsed\footnote{Probability vectors $\{ \{
  \boldsymbol{\phi}_{n}^{(p)}\}_{n=1}^{N^{(p)}(T)} \}_{p=1}^P$ can be
  integrated out using Dirichlet--multinomial conjugacy.}
slice-with-Gibbs algorithm that sequentially samples each parameter
from its conditional posterior. Pseudocode for this approach is given
in algorithm~\ref{alg:bec}. Each parameter is sampled in a univariate
fashion, except for $\boldsymbol{\beta}^{(p)}$, which is drawn using
multivariate slice sampling with the hyperractangle
method~\citep{Neal2003}. To improve mixing, we drew ten samples of
$\boldsymbol{\beta}^{(p)}$ during each Gibbs sweep. When implemented
in Python, we were able to draw 4,000 posterior samples (including
1,000 burn-in samples) of $\{ \alpha^{(p)}, \boldsymbol{\beta}^{(p)},
\{ \rho^{(qp)} \}_{q \neq p}, \tau_L^{(p)} \}_{p=1}^P$ in at most a
couple of hours for all data sets used in our experiments.

\begin{algorithm}
\caption{Inference Algorithm}
\label{alg:bec}
\begin{algorithmic}
\For{$i=1,2,\cdots,I$}
\For{$p=1,2,\cdots,P$}
\State Slice sample $\alpha^{(p)}$
\State Slice sample $\tau^{(p)}_L$
\For {$q \neq p$}
\State Slice sample $\rho^{(qp)}$
\EndFor
\For{$j=1,2,\cdots,10$}
\State Slice sample $ \boldsymbol{\beta}^{(p)}$ (multivariate)
\EndFor
\EndFor
\EndFor
\end{algorithmic}
\end{algorithm}

\section{EXPERIMENTS}

The salient characteristics of all data sets used in our experiments
are provided in table~\ref{tab:data-salient}. For each data set
obtained from TalkBank~\citep{macwhinney2007talkbank}, the ``TalkBank''
column contains the data set identifier within the ``Meetings''
section of the TalkBank database. The ``No. Tokens'' column indicates
the total number of tokens in each data set after restricting the
vocabulary to the $V=600$ most frequent stemmed types. The ``Tokens
Removed'' column contains the percentage of tokens that were discarded
via this step.

\begin{table*}[!htb]
\caption{Salient Characteristics of Data Sets.} \label{tab:data-salient}
\scriptsize
\begin{center}
\begin{tabular}{@{}llcccc@{}}
\toprule
Data Set                      & TalkBank & No.~People & No.~Utterances & No.~Tokens & Tokens Removed \\ \midrule
Synthetic                    & --                 & 3                  & 300               & 15,070        & 0.00\%         \\
University Lecture           & SB/12            & 5                  & 138               & 3,482         & 4.42\%         \\
Birthday Party               & SB/49            & 8                  & 454               & 4,229         & 5.88\%         \\
DC v. Heller                 & SCOTUS/07-290    & 10                 & 365               & 15,104        & 7.21\%         \\
L\&G v. Texas                & SCOTUS/02-102    & 6                  & 200               & 8,573         & 5.47\%         \\
Citizens United v. FEC       & SCOTUS/08-205b   & 10                 & 345               & 12,700        & 7.41\%         \\
12 Angry Men                 &  --                & 12                 & 312               & 6,350         & 5.25\%         \\
January 29, 2008 FOMC Meeting & --  & 4                  & 101               & 13,505        & 13.74\%        \\ \bottomrule
\end{tabular}
\end{center}
\end{table*}

Table~\ref{tab:log_prob_supp} contains predictive log probabilities for
several additional data sets. The ``Family Discussion'' and
``University Lecture'' data sets are conversation transcripts from the
Santa Barbara Corpus of Spoken American
English~\citep{macwhinney2007talkbank}. These data sets capture the
back-and-forth of real-world conversations. The ``January 29,
2008 FOMC Meeting'' data set is one of the FOMC meeting transcripts
used our exploratory analysis. The salient characteristics of these
data sets are given in table~\ref{tab:data-salient}. For all but one
of these additional data sets, the Bayesian Echo Chamber out-performed
a unigram language model and Blei and Lafferty's dynamic topic
model~\citeyearpar{blei06dynamic} by predicting higher probabilities
of held-out data for both a 90\%--10\% and an 80\%--20\%
training--testing split.

\begin{table*}[!htb]
  \caption{Additional Predictive Log Probabilities of Held-Out Data.}
  \label{tab:log_prob_supp}
  \scriptsize
  \centering
\begin{tabular}{@{}lr@{$\pm$}lr@{$\pm$}lrr@{$\pm$}lr@{$\pm$}lr@{}}
\toprule
& \multicolumn{5}{c}{10\% Test Set}                                         & \multicolumn{5}{c}{20\% Test Set}                                   \\  \cmidrule(l){2-11}
Data Set                               & \multicolumn{2}{c}{Our Model}  & \multicolumn{2}{c}{Unigram} & DTM~~           & \multicolumn{2}{c}{Our Model}   & \multicolumn{2}{c}{Unigram} & DTM~~     \\
\midrule
University Lecture             & -528.23           & 0.06 & -541.23        & 0.05       & \textbf{-520.74} & \textbf{-1972.67}  & 0.13 & -2009.62        & 0.12      & -2110.66  \\
Birthday Party                 & \textbf{-1883.45} & 0.11 & -1961.4        & 0.11       & -1900.68         & \textbf{-4384.42}  & 0.16 & -4625.57        & 0.20      & -4498.467 \\
January 29, 2008 FOMC Meeting & \textbf{-3187.73} & 0.04 & -3338.59       & 0.10        & -3211.09         & \textbf{-17342.43} & 0.21 & -17779.01       & 0.24      & -17726.64 \\ \bottomrule
\end{tabular}
  \end{table*}

\begin{figure*}[!htb]
\centering
  \subfloat[]{
    \includegraphics[width=0.33\textwidth]{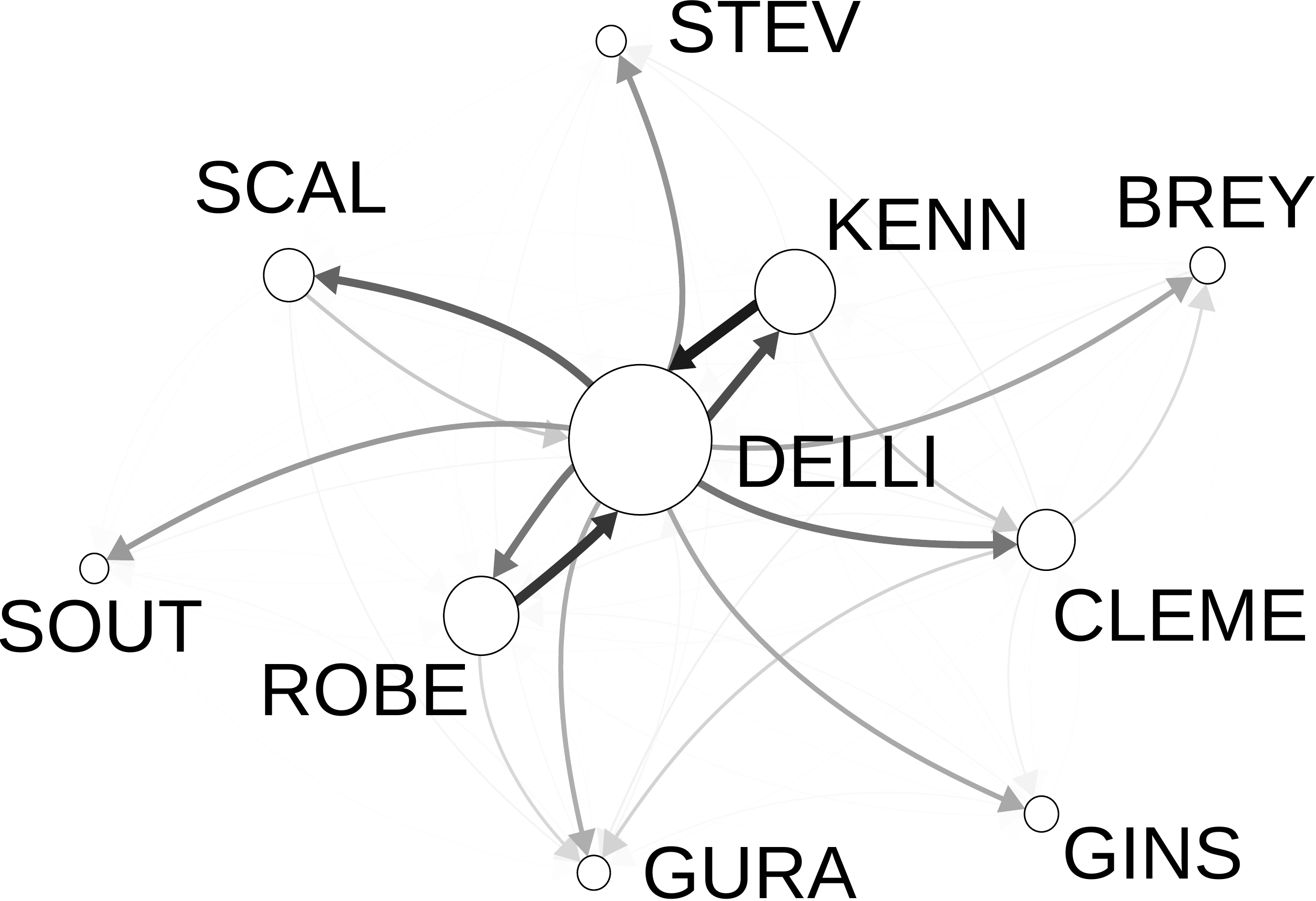}
    }
  \subfloat[]{
    \includegraphics[width=0.33\textwidth]{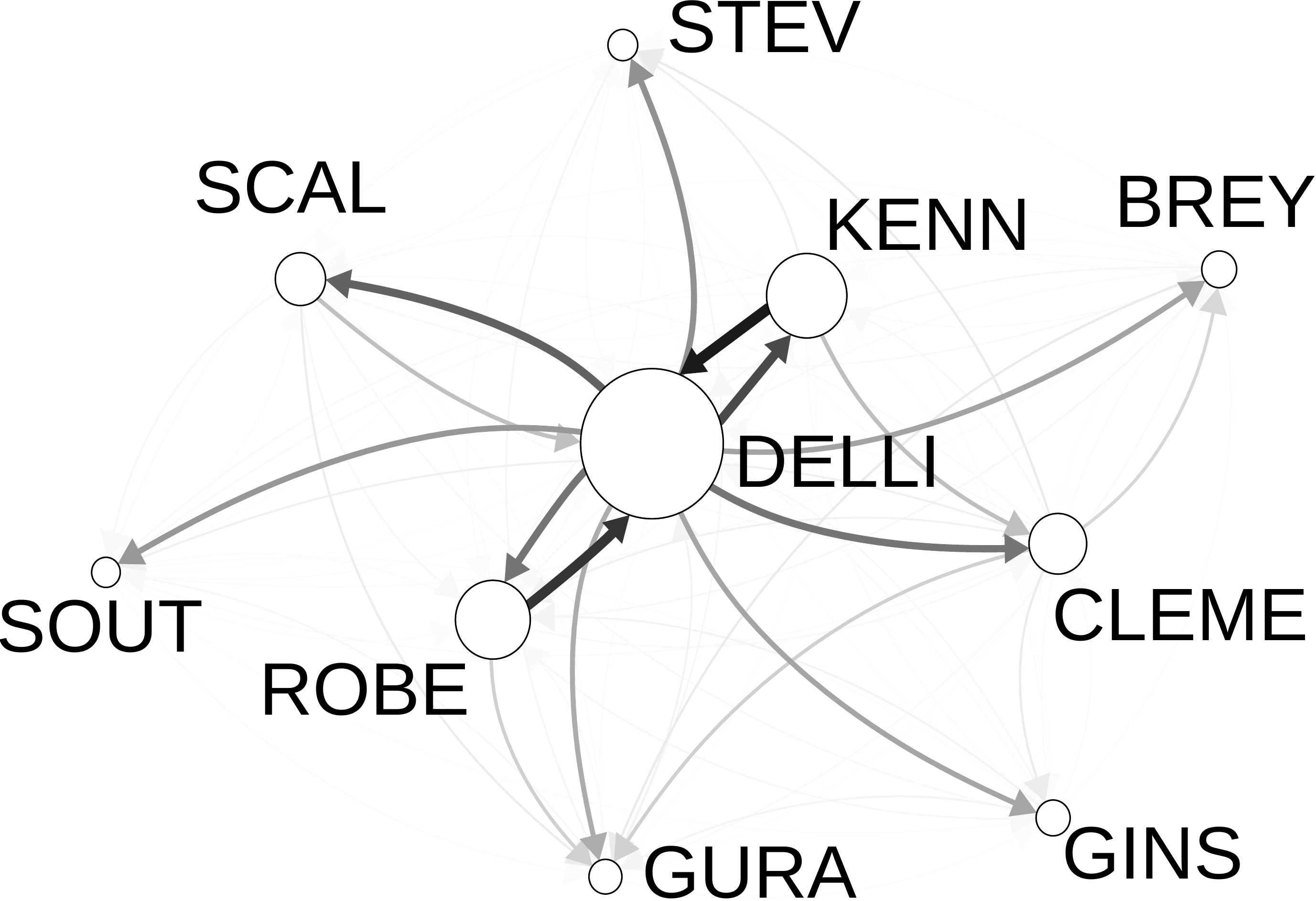}
    }
  \subfloat[]{
    \includegraphics[width=0.33\textwidth]{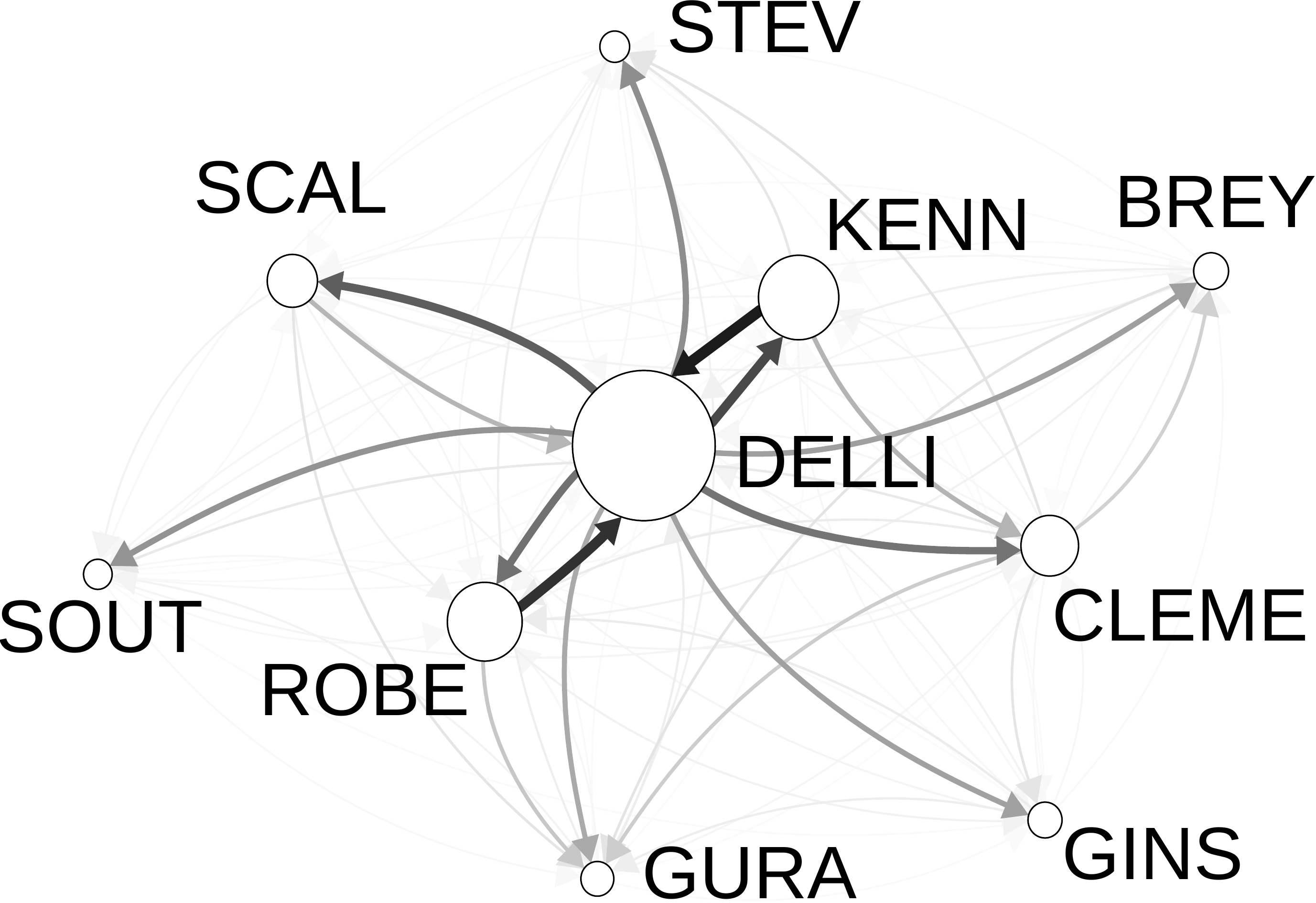}
    }
\caption{Influence Networks for the DC v. Heller Case Drawn Using (a)
  25\%, (b) 50\%, and (c) 75\% Quantiles.}
\label{fig:DcVsHeller}
\end{figure*}

Posterior means and standard deviations of the influence parameters
$\{\{\rho^{(qp)} \}_{q \neq p}\}_{p=1}^{P}$ inferred from the DC
v. Heller Supreme Court case using our model are given in
tables~\ref{tab:mat-mean-dc} and~\ref{tab:mat-sd-dc},
respectively. These values were obtained using 3,000 samples from the
posterior distribution. To further illustrate posterior uncertainty,
influence networks drawn using 25\%, 50\% (i.e., median), and 75\% posterior
quantiles are shown in figure~\ref{fig:DcVsHeller}. These networks
look very similar to each other.

\begin{table*}[!htb]
\caption{Posterior Means of $\{\{\rho^{(qp)}\}_{q \neq p} \}_{p=1}^P$
  Inferred from the DC v. Heller Case.}
\label{tab:mat-mean-dc}
\tiny
\centering
\begin{tabular}{lrrrrrrrrrr}
  \toprule
  & \multicolumn{10}{c}{To} \\
  \cmidrule(l){2-11}
  From
& DELLI & GURA & ROBE & CLEME & STEV & SCAL & KENN & GINS & SOUT & BREY \\
  \midrule
DELLI & -- & 65.85 & 109.92 & 109.36 & 86.78 & 125.29 & 143.29 & 71.21 & 82.77 & 72.98 \\
  GURA & 10.18 & -- & 4.79 & 2.43 & 7.75 & 3.27 & 2.62 & 3.17 & 6.84 & 5.08 \\
  ROBE & 161.29 & 37.99 & -- & 6.44 & 3.76 & 5.12 & 4.77 & 7.62 & 3.99 & 7.67 \\
  CLEME & 5.12 & 37.41 & 11.02 & -- & 16.53 & 4.84 & 6.03 & 15.77 & 12.53 & 32.13 \\
  STEV & 3.93 & 9.27 & 3.89 & 4.37 & -- & 3.10 & 2.70 & 2.91 & 4.42 & 3.12 \\
  SCAL & 50.53 & 15.45 & 7.77 & 5.62 & 4.64 & -- & 3.41 & 6.04 & 7.17 & 6.83 \\
  KENN & 180.91 & 2.90 & 5.86 & 50.67 & 13.75 & 4.93 & -- & 4.50 & 5.53 & 5.63 \\
  GINS & 6.98 & 9.91 & 11.29 & 4.55 & 3.22 & 4.08 & 2.69 & -- & 2.95 & 3.68 \\
  SOUT & 3.34 & 4.34 & 3.86 & 5.90 & 3.55 & 3.54 & 2.59 & 3.22 & -- & 4.50 \\
  BREY & 8.24 & 16.48 & 5.18 & 2.45 & 3.71 & 3.22 & 2.99 & 4.31 & 5.26 & -- \\
   \bottomrule
\end{tabular}
\end{table*}

\begin{table*}[!htb]
\caption{Posterior Standard Deviations of $\{\{\rho^{(qp)}\}_{q \neq p}
  \}_{p=1}^P$ Inferred from the DC v. Heller Case.}
\label{tab:mat-sd-dc}
\tiny
\centering
\begin{tabular}{lrrrrrrrrrr}
  \hline
  \toprule
  & \multicolumn{10}{c}{To} \\
  \cmidrule(l){2-11}
  From
& DELLI & GURA & ROBE & CLEME & STEV & SCAL & KENN & GINS & SOUT & BREY \\
\midrule
DELLI & -- & 7.36 & 11.08 & 8.99 & 10.30 & 12.25 & 12.73 & 10.12 & 10.94 & 9.22 \\
  GURA & 6.12 & -- & 3.88 & 2.34 & 5.90 & 3.07 & 2.51 & 3.12 & 5.20 & 4.39 \\
  ROBE & 14.60 & 12.93 & -- & 5.76 & 3.48 & 4.92 & 4.75 & 6.68 & 3.91 & 7.48 \\
  CLEME & 4.69 & 6.32 & 7.11 & -- & 9.73 & 4.45 & 5.15 & 9.35 & 8.49 & 8.81 \\
  STEV & 3.65 & 7.56 & 3.68 & 4.16 & -- & 3.12 & 2.73 & 2.99 & 4.40 & 3.08 \\
  SCAL & 14.02 & 9.42 & 6.84 & 5.02 & 4.43 & -- & 3.33 & 5.67 & 6.46 & 6.28 \\
  KENN & 14.84 & 2.70 & 5.46 & 17.43 & 10.61 & 4.68 & -- & 4.14 & 5.13 & 5.64 \\
  GINS & 6.32 & 8.14 & 9.74 & 4.24 & 3.09 & 4.09 & 2.58 & -- & 2.79 & 3.56 \\
  SOUT & 3.52 & 4.11 & 3.75 & 5.85 & 3.72 & 3.25 & 2.47 & 3.16 & -- & 4.86 \\
  BREY & 7.19 & 8.40 & 4.89 & 2.43 & 3.58 & 3.05 & 2.95 & 3.91 & 4.53 & -- \\
  \bottomrule
\end{tabular}
\end{table*}

\begin{figure*}[!ht]
\centering
  \subfloat[]{
    \includegraphics[width=0.33\textwidth]{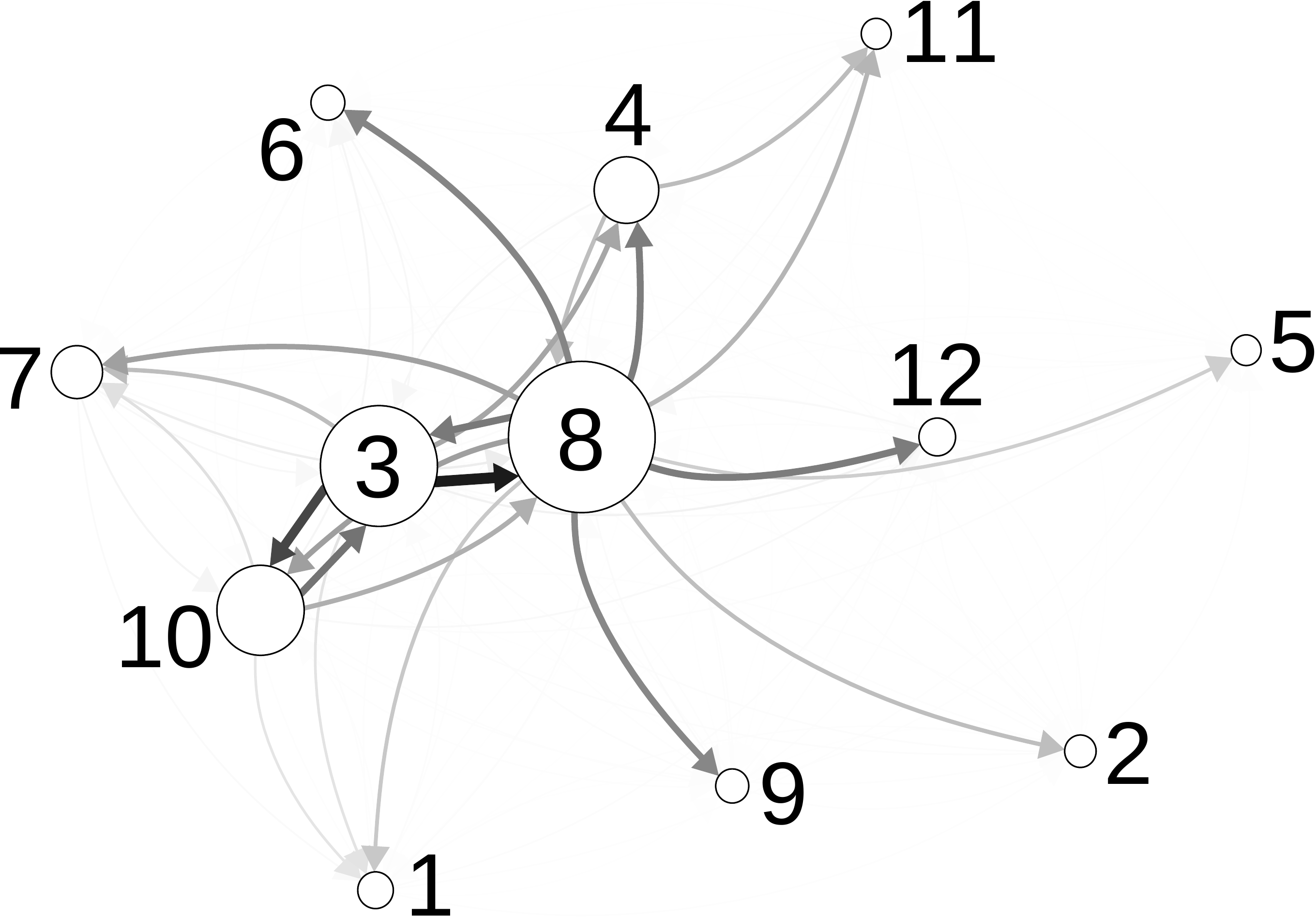}
    }
  \subfloat[]{
    \includegraphics[width=0.33\textwidth]{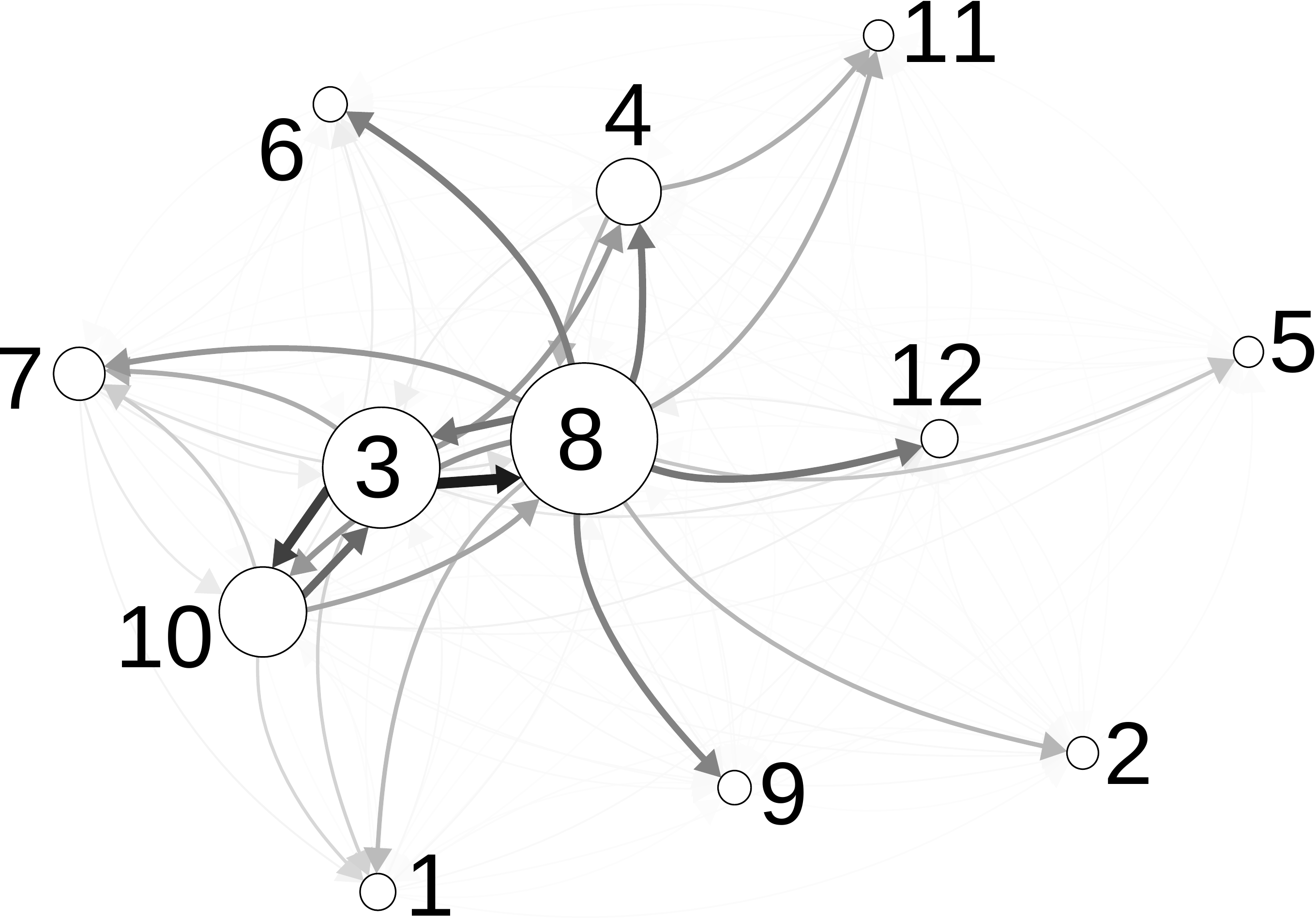}
    }
  \subfloat[]{
    \includegraphics[width=0.33\textwidth]{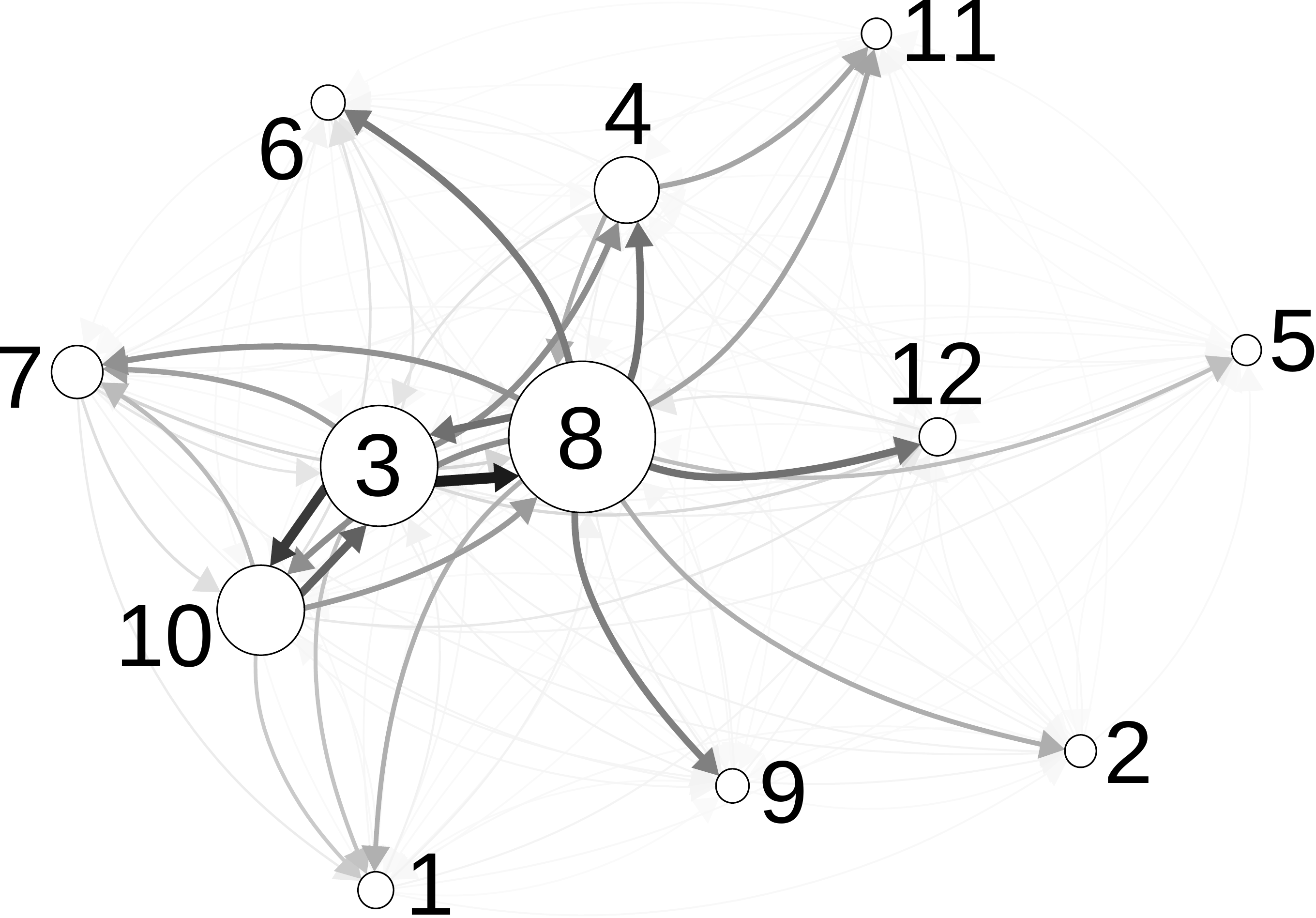}
    }
\caption{Influence Networks for ``12 Angry Men'' Drawn Using (a) 25\%,
  (b) 50\%, and (c) 75\% Quantiles.}
\label{fig:12AngryMen_supp}
\end{figure*}

Posterior means and standard deviations of the influence parameters
$\{\{\rho^{(qp)} \}_{q \neq p}\}_{p=1}^{P}$ inferred from ``12 Angry
Men'' using our model are provided in tables~\ref{tab:mat-mean-12}
and~\ref{tab:mat-sd-12}, respectively. These values were obtained
using 3,000 samples from the posterior distribution. To further
illustrate posterior uncertainty, influence networks drawn using 25\%,
50\%, and 75\% posterior quantiles are provided in
figure~\ref{fig:12AngryMen_supp}. As with the DC v. Heller case, these
networks look very similar to one another.

\begin{table*}[!htb]
\caption{Posterior Means of $\{\{\rho^{(qp)}\}_{q \neq p}\}_{p=1}^P$
  Inferred from ``12 Angry Men.''}
\label{tab:mat-mean-12}
\tiny
\centering
\begin{tabular}{lrrrrrrrrrrrr}
  \toprule
  & \multicolumn{12}{c}{To} \\
    \cmidrule(l){2-13}
      From
& Juror 8 & Juror 3 & Juror 10 & Juror 7 & Juror 1 & Juror 4 & Juror 6 & Juror 11 & Juror 12 & Juror 9 & Juror 2 & Juror 5 \\
  \midrule
Juror 8 & -- & 82.09 & 61.82 & 61.08 & 39.80 & 80.38 & 75.30 & 48.96 & 80.12 & 72.86 & 43.58 & 33.98 \\
  Juror 3 & 134.35 & -- & 112.40 & 47.76 & 27.56 & 59.83 & 10.62 & 6.46 & 16.85 & 5.28 & 5.51 & 6.84 \\
  Juror 10 & 53.38 & 88.01 & -- & 30.65 & 24.54 & 4.71 & 12.72 & 3.41 & 9.48 & 4.08 & 4.25 & 5.50 \\
  Juror 7 & 19.56 & 11.53 & 13.97 & -- & 8.24 & 3.69 & 5.19 & 3.35 & 4.46 & 4.25 & 4.86 & 4.08 \\
  Foreman & 5.12 & 5.51 & 3.30 & 2.99 & -- & 3.02 & 3.74 & 2.66 & 4.82 & 2.57 & 3.11 & 3.18 \\
  Juror 4 & 43.05 & 11.73 & 2.88 & 2.88 & 3.44 & -- & 2.47 & 46.62 & 4.08 & 6.47 & 4.55 & 3.46 \\
  Juror 6 & 5.79 & 3.23 & 3.03 & 3.16 & 2.76 & 2.56 & -- & 2.82 & 3.14 & 3.11 & 3.30 & 3.23 \\
  Juror 11 & 3.39 & 2.76 & 2.63 & 2.17 & 2.28 & 2.80 & 2.32 & -- & 2.61 & 2.50 & 2.40 & 2.44 \\
  Juror 12 & 9.61 & 3.49 & 3.00 & 3.47 & 2.84 & 2.91 & 4.44 & 3.59 & -- & 2.64 & 3.62 & 2.76 \\
  Juror 9 & 4.28 & 3.44 & 2.56 & 2.95 & 2.68 & 2.73 & 2.96 & 4.44 & 3.05 & -- & 2.67 & 2.82 \\
  Juror 2 & 2.85 & 2.99 & 2.84 & 2.67 & 3.34 & 2.41 & 3.34 & 2.16 & 3.14 & 2.49 & -- & 3.05 \\
  Juror 5 & 2.88 & 2.49 & 2.59 & 2.38 & 2.54 & 2.23 & 2.47 & 2.77 & 2.53 & 2.73 & 2.35 & -- \\
   \bottomrule
\end{tabular}
\end{table*}

\begin{table*}[!ht]
\caption{Posterior Standard Deviations of $\{\{\rho^{(qp)}\}_{q \neq p}\}_{p=1}^P$ Inferred from ``12 Angry Men.''}
\label{tab:mat-sd-12}
\tiny
\centering
\begin{tabular}{rrrrrrrrrrrrr}
  \toprule
  & \multicolumn{12}{c}{To} \\
    \cmidrule(l){2-13}
      From
& Juror 8 & Juror 3 & Juror 10 & Juror 7 & Juror 1 & Juror 4 & Juror 6 & Juror 11 & Juror 12 & Juror 9 & Juror 2 & Juror 5 \\
\midrule
Juror 8 & -- & 13.10 & 14.50 & 14.09 & 14.12 & 14.54 & 13.52 & 12.53 & 13.51 & 11.50 & 12.35 & 11.31 \\
  Juror 3 & 15.21 & -- & 18.19 & 17.49 & 16.01 & 16.63 & 8.71 & 5.76 & 12.62 & 4.86 & 5.00 & 6.17 \\
  Juror 10 & 14.47 & 16.84 & -- & 17.76 & 14.06 & 4.36 & 10.20 & 3.33 & 8.20 & 3.82 & 4.22 & 5.15 \\
  Juror 7 & 12.34 & 10.15 & 10.36 & -- & 7.30 & 3.44 & 4.91 & 3.09 & 4.66 & 4.08 & 4.57 & 3.80 \\
  Foreman & 4.58 & 5.18 & 3.42 & 3.12 & -- & 3.10 & 3.80 & 2.63 & 4.66 & 2.57 & 3.16 & 3.06 \\
  Juror 4 & 12.45 & 9.17 & 2.84 & 2.81 & 3.32 & -- & 2.44 & 15.45 & 4.04 & 6.41 & 4.27 & 3.38 \\
  Juror 6 & 5.51 & 3.42 & 3.04 & 3.14 & 2.66 & 2.44 & -- & 2.72 & 3.06 & 3.06 & 3.43 & 3.30 \\
  Juror 11 & 3.46 & 2.75 & 2.55 & 2.26 & 2.26 & 2.77 & 2.28 & -- & 2.60 & 2.47 & 2.33 & 2.41 \\
  Juror 12 & 8.14 & 3.38 & 3.00 & 3.44 & 2.76 & 2.87 & 4.24 & 3.51 & -- & 2.63 & 3.46 & 2.63 \\
  Juror 9 & 4.35 & 3.39 & 2.51 & 2.97 & 2.64 & 2.70 & 2.90 & 4.59 & 3.00 & -- & 2.58 & 2.76 \\
  Juror 2 & 2.79 & 2.90 & 2.68 & 2.64 & 3.42 & 2.23 & 3.41 & 2.20 & 3.07 & 2.49 & -- & 3.21 \\
  Juror 5 & 2.83 & 2.53 & 2.68 & 2.49 & 2.46 & 2.28 & 2.55 & 2.76 & 2.63 & 2.80 & 2.35 & -- \\
   \bottomrule
\end{tabular}
\end{table*}

Log probabilities, obtained using a 90\%--10\% training--testing split
and a vocabulary of $V=300$ types, are provided for the tied and untied
combined models in table~\ref{tab:tiedVSuntied}.  The tied
model, whose likelihood is the product of the Bayesian Echo Chamber's
likelihood and that of Blundell et al.'s model but with shared
influence parameters, assigned lower probabilities to held-out data
than the fully factorized (i.e., untied) model.

\begin{table}[!htb]
  \scriptsize
  \centering
\begin{tabular}{lr@{$\pm$}l@{ }r@{$\pm$}l}
\toprule
Data Set & \multicolumn{2}{c}{Tied} & \multicolumn{2}{c}{Untied}
\\ \midrule
L\&G v.~Texas & -5507.11 & 0.15 & \textbf{-5502.87} & 0.15
\\
DC v.~Heller & -6321.30 & 0.16 & \textbf{-6303.55} & 0.15
\\
Citizens United v.~FEC & -4795.24 & 0.18 & \textbf{-4777.96} & 0.17
\\
``12 Angry Men'' & -4014.56 & 0.24 & \textbf{-3987.20} & 0.23
\\ \bottomrule
\end{tabular}
\caption{Log Probabilities of Held-Out Data for the
  Combined Model with Tied and Untied Parameters.}
\label{tab:tiedVSuntied}
\end{table}

\subsubsection*{Acknowledgements}

Thanks to Juston Moore and Aaron Schein for their work on early stages
of this project, and to Aaron for the ``Bayesian Echo Chamber''
model name. This work was supported in part by the Center for Intelligent
Information Retrieval, in part by NSF grant \#IIS-1320219, and in part
by NSF grant \#SBE-0965436. Any opinions, findings and conclusions or
recommendations expressed in this material are the authors' and do not
necessarily reflect those of the sponsor.


\bibliographystyle{apalike}
\bibliography{refs}

\end{document}